\documentclass[sigconf,nonacm]{acmart}

\newcommand{\zqy}{\color{black}}
\newcommand{\cheng}{\color{black}}
\newcommand{\qy}{\color{black}}

\newcommand{\qycikm}{\color{black}}
\newcommand{\zqycikm}{\color{black}}
\newcommand{\zlcikm}{\color{black}}

\usepackage{graphicx}
\usepackage{enumitem}
\usepackage{subcaption}
\usepackage{multirow}

\setcopyright{none}
\begin{document}

\title{HHGT: Hierarchical Heterogeneous Graph Transformer for Heterogeneous Graph
Representation Learning}

\author{Qiuyu Zhu}
\affiliation{%
  \institution{Nanyang Technological University}
  \country{Singapore}
}
\email{qiuyu002@e.ntu.edu.sg}

\author{Liang Zhang}
\affiliation{%
  \institution{Nanyang Technological University}
  \country{Singapore}
}
\authornotemark[2]
\email{liang012@e.ntu.edu.sg}

\author{Qianxiong Xu}
\affiliation{%
  \institution{Nanyang Technological University}
  \country{Singapore}
}
\email{qianxion001@e.ntu.edu.sg}

\author{Kaijun Liu}
\affiliation{%
  \institution{Nanyang Technological University}
  \country{Singapore}
}
\email{kaijun001@e.ntu.edu.sg}

\author{Cheng Long}
\affiliation{%
  \institution{Nanyang Technological University}
  \country{Singapore}
}
\email{c.long@ntu.edu.sg}
\authornotemark[2]

\author{Xiaoyang Wang}
\affiliation{%
  \institution{University of New South Wales}
  \country{Australia}
}
\email{xiaoyang.wang1@unsw.edu.au}
\authornotemark[2]





\begin{abstract}
Despite the success of Heterogeneous Graph Neural Networks (HGNNs) in modeling real-world Heterogeneous {\qycikm Information Networks} (HINs), challenges such as expressiveness limitations and over-smoothing have prompted researchers to explore Graph Transformers (GTs) for enhanced {\qycikm HIN} representation learning. 
However, research on GT in HINs {\zlcikm remains} limited, with two key {\zlcikm shortcomings} in existing work:
(1) A node's neighbors at different distances in HINs convey diverse semantics; for instance, a paper's {\zlcikm direct} neighbor (a paper) in an academic graph signifies a citation relation, whereas the indirect neighbor (another paper) implies a thematic association, reflecting distinct meanings. Unfortunately, existing methods ignore such differences and uniformly treat neighbors within a given distance in a coarse manner, which results in semantic confusion.
(2) Nodes in HINs have various types, each with unique semantics, e.g., papers and authors in {\zlcikm an} academic graph carry distinct meanings. Nevertheless, existing methods mix nodes of different types during neighbor aggregation, {\zlcikm hindering the capture of proper correlations between nodes of diverse types.}
To bridge these gaps, we design an innovative structure named {\zlcikm $(k,t)$-ring neighborhood, where nodes are initially organized by their distance, forming different non-overlapping $k$-ring neighborhoods for each distance. Within each $k$-ring structure, nodes are further categorized into different groups according to their types, thus emphasizing the heterogeneity of both distances and types in HINs naturally.} Based on this structure, we propose a novel \textbf{\underline{H}}ierarchical \textbf{\underline{H}}eterogeneous \textbf{\underline{G}}raph \textbf{\underline{T}}ransformer (HHGT) model, which seamlessly integrates a Type-level Transformer for aggregating nodes of different types within each $k$-ring neighborhood, followed by a Ring-level Transformer for aggregating different $k$-ring neighborhoods in a hierarchical manner. {\qycikm Extensive experiments are conducted on downstream tasks to verify HHGT's superiority over {\zlcikm \textbf{14}} baselines, with a notable improvement of up to 24.75\% in NMI and 29.25\% in ARI for node clustering {\zlcikm task} on the ACM dataset compared to the best baseline.}

\end{abstract}








\maketitle

\section{Introduction}
\label{sec:intro}

\begin{figure}[th]
    \centering
    \includegraphics[width=0.49\textwidth]{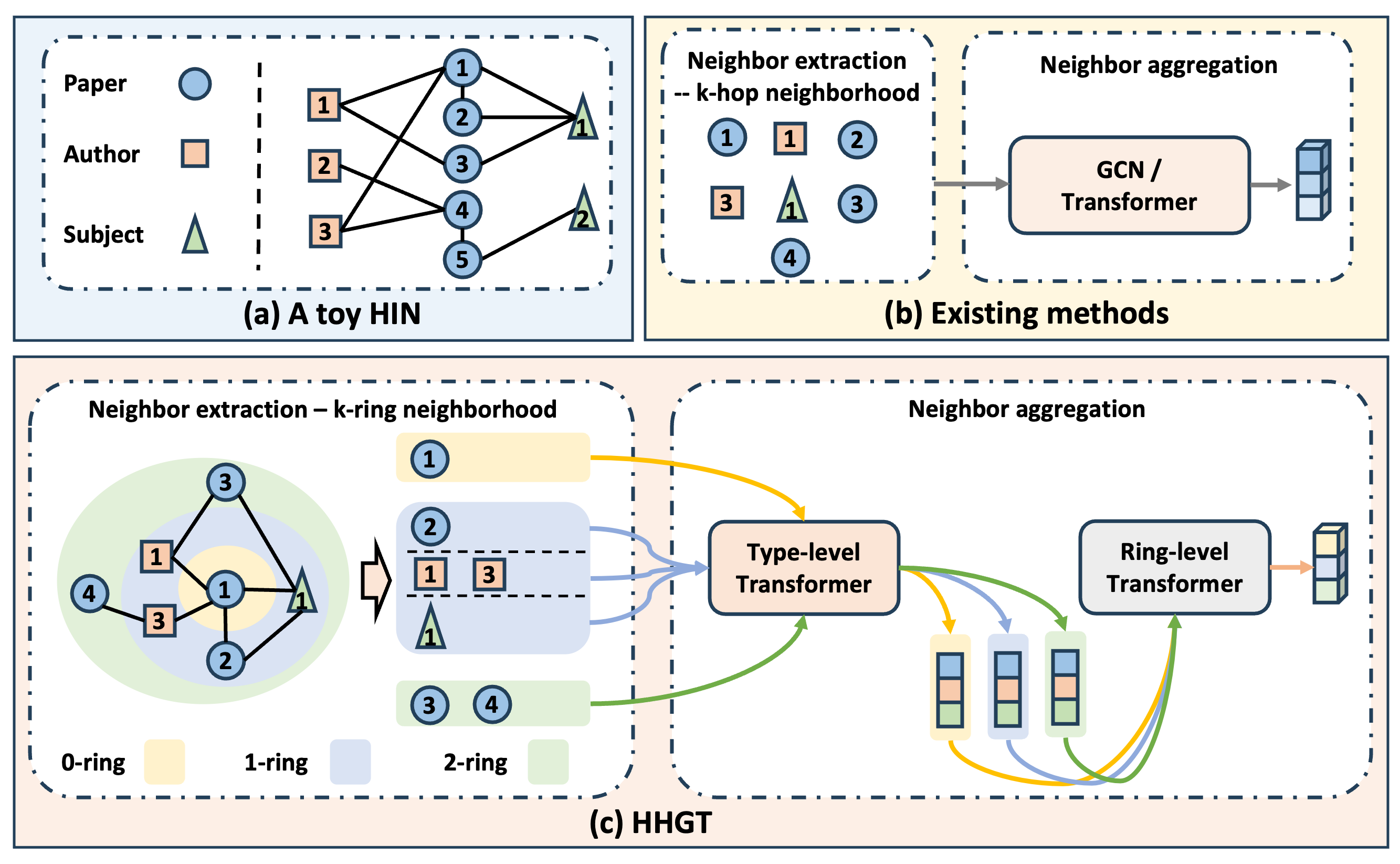}
    \caption{The difference between existing HGT-based methods and our HHGT model: (a) A toy HIN; (b) Learning node representation through existing HGT-based methods; (c) Learning node representation through our HHGT model. Here, node {\qycikm $P_1$'s} $0$-ring neighborhood, $1$-ring neighborhood, $2$-ring neighborhood are visually highlighted with yellow, blue, green colors, respectively. Within $1$-ring neighborhood, nodes of different types are separated by dashed lines.}
    \label{fig:intro}
\end{figure}

Heterogeneous Information Networks (HINs)~\cite{sun2013mining}, also well-known as Heterogeneous Graphs (HGs), consist of multiple types of objects (i.e., nodes) and relations (i.e., edges).
They are prevalent in real-world scenarios, ranging from citation networks~\cite{hamilton2017inductive,wang2016structural}, social networks~\cite{atwood2016diffusion,kipf2016semi} to recommendation systems~\cite{berg2017graph,zhang2019star}.
For example, the academic data shown in Figure~\ref{fig:intro}(a) can be represented as an HIN, which contains three types of nodes (i.e., paper, author, subject) and three types of relations (i.e., author-write-paper, paper-belong-subject, paper-cite-paper). Recently, there has been a notable surge in research focusing on representation learning for HINs~\cite{dong2017metapath2vec,fu2017hin2vec,shi2018heterogeneous,shi2018aspem}, which emerges as a powerful technique for embedding nodes into low-dimensional representations while retaining both graph structures and heterogeneity. 

Given the success of traditional Graph Neural Networks (GNNs) \cite{kipf2016semi,vaswani2017attention,hamilton2017inductive} in handling homogeneous graphs (containing {\qycikm only} one type of nodes and relations), researchers are increasingly turning their attention to HIN representation learning using GNNs, known as Heterogeneous Graph Neural Networks (HGNNs). HGNN-based approaches \cite{fu2020magnn,liu2022aspect} often leverage neighbor aggregation strategies to effectively capture and propagate information across diverse types of nodes in HINs. For example, R-GCN~\cite{schlichtkrull2018modeling} extends the traditional Graph Convolutional Networks (GCNs) ~\cite{kipf2016semi} by incorporating relation-specific weight matrices, aiming to capture the diverse relations within an HIN. Fu et al.~\cite{fu2020magnn} propose to incorporate intermediate nodes along meta-paths, using both intra-meta-path and inter-meta-path information for higher-order semantic information aggregation.

Despite HGNNs have achieved success in modeling real-world HINs, the presence of challenges such as limitations in expressiveness~\cite{xu2018powerful}, over-smoothing~\cite{chen2020measuring} and over-squashing~\cite{alon2020bottleneck} has driven researchers to investigate Graph Transformers (GTs)~\cite{ying2021transformers} for enhanced HIN representation learning. For instance, Hu et al.~\cite{hu2020heterogeneous} propose a heterogeneous Transformer-like attention architecture for neighbor aggregation. Mao et al.~\cite{mao2023hinormer} leverage a local structure encoder and a heterogeneous relation encoder to capture structure and heterogeneity information in HINs. In general, existing GT-based methodologies for HIN representation learning, i.e., HGT-based methods, depicted in Figure~\ref{fig:intro}(b), adhere to a typical principle:
Given a target node, its $k$-hop neighborhood (i.e., those nodes within a reachable distance of $\leq k$ from the target node) is first extracted.
Then, GCN~\cite{kipf2016semi} or Transformer~\cite{vaswani2017attention} would be utilized to propagate information from these nodes to the target node.

Nevertheless, existing HGT-based approaches tend to mix nodes of different types and uniformly treat all nodes within $k$-hop neighborhood during neighbor aggregation, leading to potential semantic confusion. In particular, (1) \textbf{Limitation 1:} {\zlcikm Neighbors of a target node at different distances in HINs carry varied semantics.} Using Figure~\ref{fig:intro}(a) as an illustration, paper {\qycikm $P_1$'s} direct neighbor, paper {\qycikm $P_2$}, indicates a citation relation. Conversely, the indirect neighbor, paper {\qycikm $P_3$}, implies a thematic connection without a direct citation relation, showcasing different connotations. {\qycikm Regrettably, existing strategies overlook such distinctions by uniformly addressing each neighbor within distance $k$, i.e., packing $P_1, P_2, P_3$ together into a single sequence and aggregating them uniformly. This is not desirable since these nodes serve different functions.}
{\qycikm (2) \textbf{Limitation 2:} {\zlcikm Neighbors of a target node with different types also carry distinct semantics.} Taking Figure~\ref{fig:intro}(a) as an instance, paper {\qycikm $P_1$'s} direct neighbors include paper {\qycikm $P_2$}, author $A_1,A_3$ and subject $S_1$. {\zlcikm Here, {\qycikm $P_2$} represents a citation relation, $A_1,A_3$ reflect authorship relations, while $S_1$ signifies a topic alignment relation.}}
While existing HGT-based methods consider node types, they typically pack {\qycikm $P_2,A_1,A_3,S_1$} together as a unified sequence. {\zlcikm This approach is not desirable because it mixes nodes of different types during neighbor aggregation, blurring the distinct functions of papers, authors, and subjects.}

To overcome these challenges, we propose the following two main designs: (1) \textbf{Design 1:} 
{\qycikm To distinguish a node's neighbors at varying distances, we introduce an innovative structure called the $k$-ring neighborhood. This structure specifically refers to nodes whose distance from the target node is exactly $k$, differentiating it from the commonly known 
$k$-hop neighborhood.} In essence, we split the $k$-hop neighborhood into $k+1$ non-overlapping $k$-ring neighborhoods, where the nodes in each $k$-ring neighborhood share the same distance to the target node. As illustrated in Figure~\ref{fig:intro}(c), considering $k=2$, for paper {\qycikm $P_1$}, its neighbors within a distance of $2$ can be decomposed into three distinct $k$-ring neighborhoods: the $0$-ring neighborhood {\qycikm $\{P_1\}$}, the $1$-ring neighborhood {\qycikm $\{P_2,A_1,A_3,S_1\}$}, and the $2$-ring neighborhood {\qycikm $\{P_3,P_4\}$}. Building upon this new structure, we extract diverse $k$-ring neighborhoods for each node, {\qycikm which can naturally discern different functions and thus preventing semantic confusion. Then, a Ring-level Transformer is designed to aggregate distinct $k$-ring neighborhoods separately, with aggregation based on the relevance and significance of each $k$-ring neighborhood to the target node.} (2) \textbf{Design 2:} {\qycikm To avoid mixing nodes of different types within each $k$-ring structure, we further propose a novel $(k,t)$-ring structure by arranging nodes into different groups based on their types within each $k$-ring structure.} Based on such neighborhood partition, a Type-level Transformer is proposed to separately aggregate neighbors of distinct types for a target node within each $k$-ring structure, considering the importance of each type to the target node. In Figure~\ref{fig:intro}(a), consider the $1$-ring neighborhood of node {\qycikm $P_1$} (i.e., {\qycikm ${P_2, A_1, A_3, S_1}$}), where nodes of diverse types coexist. We partition this $1$-ring neighborhood into three groups based on node types, namely, paper {\qycikm $P_2$}, author $A_1, A_3$, and subject $S_1$, with each group carrying unique {\zlcikm functions. Then, we apply a Type-level Transformer to aggregate each group separately, rather than treating them as a unified sequence as done by existing HGT-based methods. This approach enables us to mimic the diverse roles of nodes with various types.}

{\zlcikm In summary, for each target node, we extract its neighbors from diverse $k$-ring neighborhoods, where the nodes within each ring are further grouped according to their types, forming an innovative $(k,t)$-ring neighborhood structure.
Building upon this structure}, we introduce a novel \textbf{\underline{H}}ierarchical \textbf{\underline{H}}eterogeneous \textbf{\underline{G}}raph \textbf{\underline{T}}ransformer (HHGT) model. This model seamlessly integrates a Type-level Transformer for aggregating nodes of different types within each $k$-ring neighborhood separately, followed by a Ring-level Transformer for aggregating different $k$-ring neighborhoods in a hierarchical manner. 
The main contributions of our paper are summarized as follows:
\vspace{-1.5mm}
\begin{itemize}[leftmargin=*]
\item {\qycikm For the first time, we design an innovative $(k,t)$-ring neighborhood structure for HIN representation learning, which emphasizes the heterogeneity of both distances and types in HINs naturally.}
\item To the best of our knowledge, we are the first to propose a hierarchical graph transformer model for node representation learning in HINs, which seamlessly integrates a Type-level Transformer for aggregating nodes of distinct types within each $k$-ring structure separately, followed by hierarchical aggregation utilizing a Ring-level Transformer for different $k$-ring neighborhoods.
\item {\zlcikm Extensive experimental results on two real-world HIN benchmark datasets demonstrate that our model significantly outperforms \textbf{14} baseline methods on  two typical downstream tasks.
Additionally, the ablation study validates the advantages and significance of considering the heterogeneity of both distances and types in HINs.}
\end{itemize}
\vspace{-2mm}

\section{Related Work}
\label{sec:related}

\subsection{Shallow Models for HIN Embedding} 
In recent years, a plethora of graph embedding techniques~\cite{grover2016node2vec,hamilton2017inductive} have emerged with the goal of mapping nodes or substructures into a low-dimensional space, preserving the connecting structures within the graph. 
As real-world networks typically consist of various types of nodes and relations~\cite{sun2013mining}, research on shallow models for HIN embedding~\cite{shi2016survey,wang2022survey,yang2020heterogeneous} has garnered significant attention. Shallow models for HIN embedding can be broadly classified into random walk-based  methods~\cite{dong2017metapath2vec,shi2018heterogeneous} and first/second-order proximity-based methods~\cite{tang2015pte,fu2017hin2vec,shi2018aspem,zhang2018metagraph2vec}. For instance, Metapath2vec~\cite{dong2017metapath2vec} adopts meta-path guided random walk to acquire the semantic information between pairs of nodes. These methods leverage meta-paths or type-aware network closeness constraints to exploit network heterogeneity for HIN embedding. Despite their contributions, these shallow models lack the ability to effectively capture intricate relations and semantics within HINs, resulting in suboptimal representation learning.

\subsection{Deep Models for HIN Embedding}
As deep learning models have shown remarkable success in capturing both structural and content information within homogeneous graphs~\cite{cai2018comprehensive,wu2020comprehensive,liu2023gapformer,vaswani2017attention}, the research focus extended to Gs, giving rise to deep models for HINs~\cite{yang2020heterogeneous,lv2021we}. Deep models for HIN embedding are broadly categorized into two types: meta-path-based deep models~\cite{huang2016meta,wang2019heterogeneous,fu2020magnn,yang2023simple,zheng2021heterogeneous,yun2019graph} and meta-path-free deep models~\cite{schlichtkrull2018modeling,mao2023hinormer,zhao2023exploiting,zhu2019relation,liu2022aspect,hu2020heterogeneous}. Meta-path-based deep models employ meta-paths to aggregate information from type-specific neighborhoods, offering the advantage of capturing higher-order semantic information dictated by selected meta-paths. For example, HAN~\cite{wang2019heterogeneous} leverages a hierarchical attention mechanism, which considers both node-level attention and semantic-level attention to learn the importance of nodes and meta-paths, respectively. However, these approaches require expert knowledge for meta-path selection, posing a significant impact on model performance. For meta-path-free strategies, Schlichtkrull et al.~\cite{schlichtkrull2018modeling} propose to model relational data through relation-aware graph convolutional layers, enabling robust representation learning in HINs without meth-paths. Hu et al.~\cite{hu2020heterogeneous} introduce an attention mechanism inspired by Transformers, specifically designed for neighbor aggregation. Despite eliminating handcrafted meta-paths, meta-path-free deep models exhibit two key shortcomings: (1) They mix nodes of different types during neighbor aggregation, resulting in a failure to adequately capture the correlations between nodes of different types. (2) They ignore the fact that a node’s neighbors at different distances in HINs carry distinct semantics and thus treating them uniformly during neighbor aggregation, which may lead to semantic confusion and suboptimal performance on downstream tasks.

\section{Preliminaries}
\label{sec:pre}

\subsection{Problem Definition}
\noindent \textbf{Definition 3.1. Heterogeneous Information Network~\cite{sun2013mining}.} A heterogeneous information network (HIN) is formally defined as $\mathcal{G}=\{\mathcal{V},\mathcal{E},\mathcal{C},\mathcal{R}\}$, where $\mathcal{V}$, $\mathcal{E}$, $\mathcal{C}$, $\mathcal{R}$ represent the set of nodes, edges, node types, and relation types, respectively. In {\qy an} HIN, each node is associated with a node type in $\mathcal{C}$ and each edge has its corresponding relation type in $\mathcal{R}$. Each node $v\in \mathcal{V}$ is associated with a feature vector $f\in \mathbb{R}^{d}$, where $d$ is the feature dimension. An HIN is characterized by the condition $|\mathcal{C}| + |\mathcal{R}| > 2$. 


\noindent \textbf{Definition 3.2. HIN Representation {\zqy Learning} Problem~\cite{wang2022survey}.} 
Given an HIN, we aim to learn the node embedding $z_v\in \mathbb{R}^{d}$ for each node $v$, where $d \ll |\mathcal{V}|$ denotes the embedding dimension.

After learning the node representation of HINs, we can use the embeddings obtained for many downstream tasks, including semi-supervised node classification, unsupervised node clustering, etc.

\subsection{{\cheng Transformer Encoder}} 
\label{sec:encoder}
Transformer encoder, a core component of Transformer~\cite{vaswani2017attention}, consists of multiple identical layers, each containing two main sub-modules: the Multi-Head Self-Attention (MSA) module and the Feed-Forward Network (FFN) module. Both components incorporate residual connections and Layer Normalization (LN). To simplify the explanation, we just focus on the single-head self-attention module. {\cheng Given} an input sequence $\mathcal{H} \in \mathbb{R}^{n\times d}$ where $n$ denotes {\qy the} token number and $d$ denotes the hidden dimension, MSA firstly projects it to query, key and value spaces (namely $Q, K, V$, respectively), which are written as:
\begin{equation}
    Q = \mathcal{H} W_q, 
    K = \mathcal{H} W_k, 
    V = \mathcal{H} W_v, 
\end{equation}
where $W_q\in \mathbb{R}^{d\times d_k}$, $W_k\in \mathbb{R}^{d\times d_k}$, $W_v\in \mathbb{R}^{d\times d_v}$ are learnable matrices. After that, it calculates the attention scores by taking the dot product of $Q$ and the transpose of $K$, normalized by the scaling factor $\sqrt{d_k}$:
\begin{equation}
MSA(\mathcal{H}) = Softmax(\frac{QK^T}{\sqrt{d_k}}) V.
\end{equation}
Then, the MSA output is passed through FFN with an LN and a residual connection to generate the output of the $l$-th Transformer layer as:
\begin{equation}
\begin{aligned}
\Tilde{\mathcal{H}}^{(l)} &= MSA(LN(\mathcal{H}^{(l-1)})) + \mathcal{H}^{(l-1)}, \\
\mathcal{H}^{(l)} &= FFN(LN(\Tilde{\mathcal{H}}^{(l)})) + \Tilde{\mathcal{H}}^{(l)}.
\end{aligned}
\end{equation}
Here, $l=1,\dots,L$ represents the $l$-th layer of the Transformer.

\section{Methodology}
\label{sec:model}
\begin{figure*}[th]
    \centering
     \includegraphics[width=0.82\textwidth]{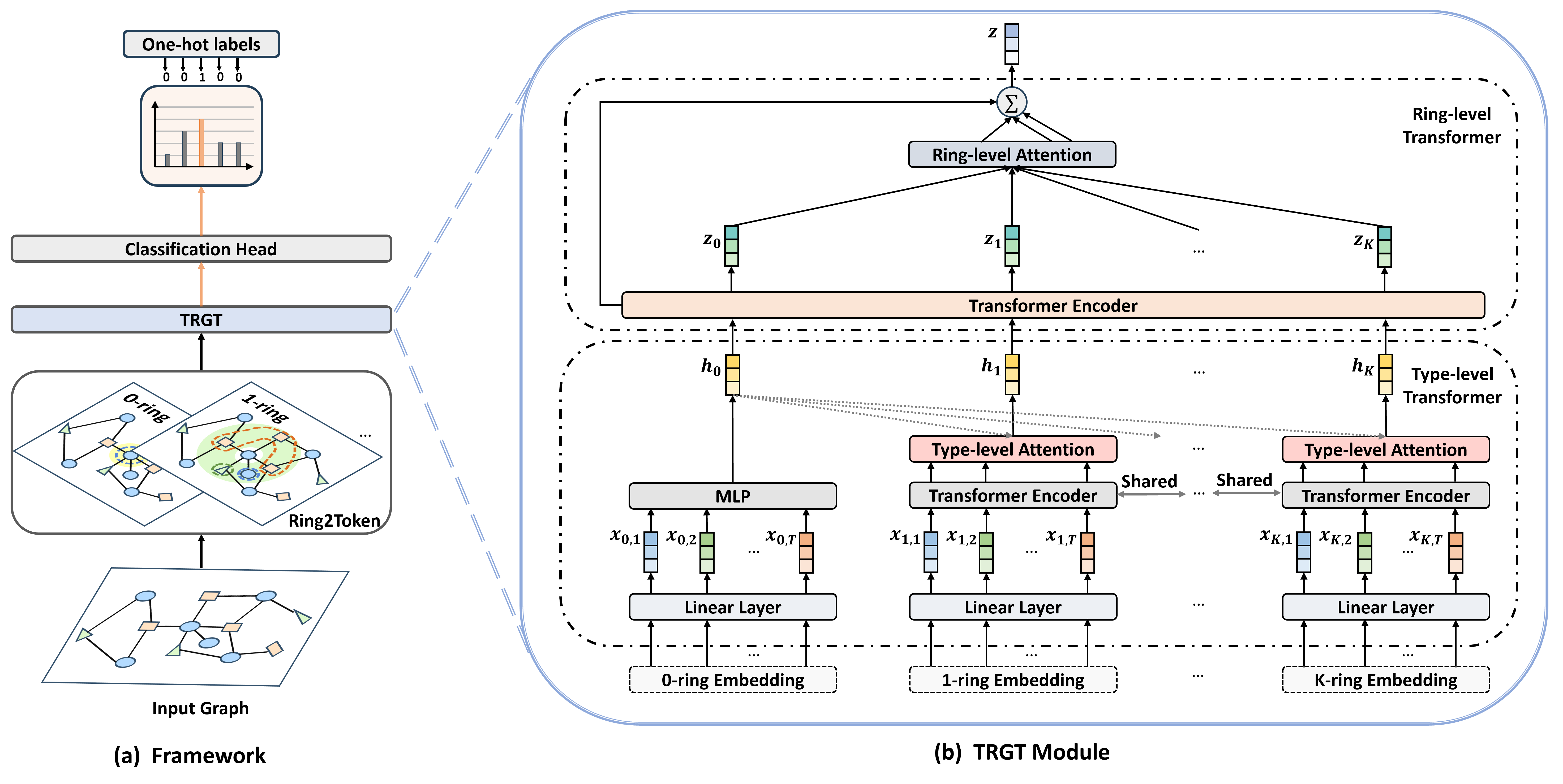}
    \caption{(a) Illustration of the framework for node classification task. (b) Diagram of the TRGT module incorporating both Ring-level Transformer and Type-level Transformer.}
    \vspace{-3mm}
    \label{fig:frame}
\end{figure*}

\begin{figure}[th]
    \centering
    \includegraphics[width=0.48\textwidth]{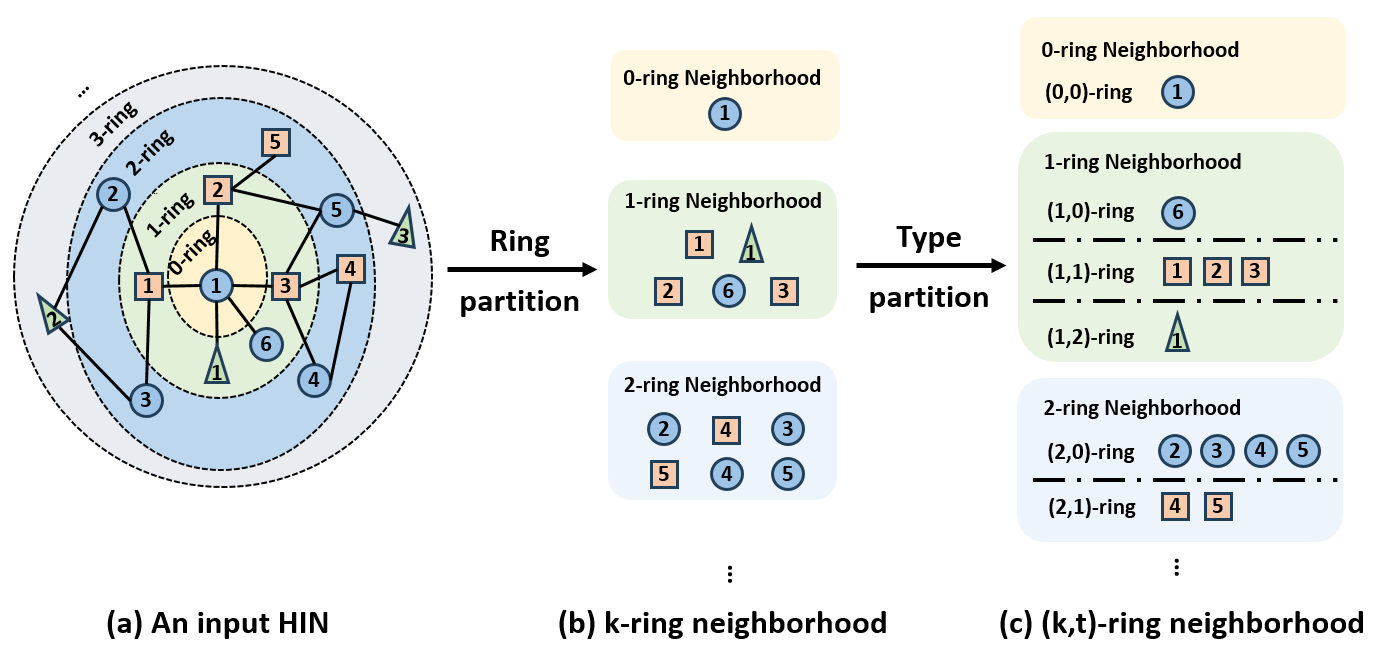}
    \caption{{\zlcikm Illustration of neighborhood partition by the Ring2Token module.}}
    \vspace{-3mm}
    \label{fig:embedding}
\end{figure}

In this section, we present the details of HHGT model, consisting of two important modules: {\qycikm Ring2Token} and TRGT. The overall framework is depicted in Figure~\ref{fig:frame}(a). Given an HIN and an integer $K$, for each target node, we initially utilize Ring2Token to extract multiple $k$-ring neighborhoods ($k\in [0,K]$), spanning from $0$-ring neighborhood to $K$-ring neighborhood, with well-organized nodes {\zlcikm partitioned} by their types within each $k$-ring structure, {\zlcikm forming the $(k,t)$-ring neighborhood structure}. 
{\zlcikm After the neighborhood partition}, we use the TRGT module to learn node representations via the GT layer {\zlcikm based on these extracted $(k,t)$-ring neighborhoods}. This {\zlcikm involves a} Type-level Transformer to aggregate nodes of different types within each $k$-ring neighborhood, followed by a Ring-level Transformer to aggregate different $k$-ring neighborhoods hierarchically. After {\zlcikm obtaining} representations for all nodes via our HHGT model, following previous work~\cite{liu2022aspect}, we apply the classification head to transform these node representations into the classification results {\zlcikm and train HHGT using the cross-entropy loss function.}
The details of Ring2Token and TRGT modules are discussed as follows.


\subsection{Ring2Token}
How to effectively aggregate information from neighbors into a node is critical for designing a powerful HIN representation learning model 
~\cite{wang2019heterogeneous,hu2020heterogeneous,fu2020magnn}. However, existing methods overlook the distinctions between neighbors at different distances and mix nodes of distinct types during neighbor aggregation.{\zlcikm To address this limitation}, we introduce Ring2Token, which considers neighbor information involving different node types at distinct distances. To grasp the distinctions between neighbors at different distances, we first design the novel $k$-ring neighborhood structure as follows.

\noindent \textbf{Definition 4.1. $k$-ring Neighborhood.} 
Given a node $u$, suppose $\Gamma_{k}(u)=\{v\in \mathcal{V} | d(u,v)=k\}$ denote the $k$-ring neighborhood of $u$, where $d(u,v)$ refers to the shortest path distance between nodes $u$ and $v$. The $0$-ring neighborhood is the target node, i.e., $\Gamma_0(u)=\{u\}$.
{\qycikm Taking Figure~\ref{fig:embedding}(a) as an example where paper $P_1$ is the target node, and paper $P_6$ is its 1-ring neighbor, while paper $P_2, P_3, P_4, P_5$ are its 2-ring neighbors. In this case, paper $P_6$ signifies a citation relation and paper $P_2, P_3, P_4, P_5$ imply thematic associations. Using the 2-hop (including $P_2, P_3, P_4, P_5, P_6$) mix nodes at different distances, thereby failing to distinguish the different functions associated with paper $P_6$ and $P_2, P_3, P_4, P_5$. In contrast, the $k$-ring separates the 2-hop neighbors into different subsets as ${P_6}$ and ${P_2, P_3, P_4, P_5}$, which naturally discerns the different functions and thus prevents semantic confusion.}

Meanwhile, each node type carries specific information, embodying distinct concepts.   
For instance, in Figure~\ref{fig:embedding}(b), the $1$-ring neighborhood of paper $P_1$ involves nodes of three different types: paper $P_6$, authors $A_1,A_2,A_3$, and subject $S_1$, where papers offer citation relations, authors contribute to the creation of the paper while subjects provide thematic information. 
{\zlcikm Therefore, it is not suitable to mix nodes of all types together during neighbor aggregation.}
Motivated by this, we introduce the concept of type-aware $k$-ring neighborhood (named $(k,t)$-ring neighborhood) to categorize nodes within each $k$-ring structure by their types.

\noindent \textbf{Definition 4.2. $(k,t)$-ring Neighborhood.} 
Given a node $u$, let $\Gamma_{k,t}(u)=\{v\in \mathcal{V} | v\in \Gamma_{k}(u) \wedge \mathcal{C}_v = t\}$ denote the $(k,t)$-ring neighborhood of $u$, where $\Gamma_{k}(u)$ refers to $u$'s $k$-ring neighborhood and $\mathcal{C}_v$ denotes $v$'s node type. 

{\qycikm Based on the concept of $(k,t)$-ring, we can further {\zlcikm partition} the 1-ring neighborhood of $P_1$ into three different subsets as $\Gamma_{1,1}(P_1)=\{P_6\}$, $\Gamma_{1,2}(P_1)=\{A_1, A_2, A_3\}$, $\Gamma_{1,3}(P_1)=\{S_1\}$. 
{\zlcikm Then, we can aggregate them separately, enabling us to mimic their distinct roles and avoid mixing different types.}}

{\qycikm To sum up,} given an HIN, for each node, {\zlcikm Ring2Token extracts and partitions all its neighborhoods with a $(k,t)$-ring structure.} Specifically, given an integer $K$ and node type number $T$, for a node $u$, it possesses a sequence of $k$-ring neighborhoods with length $K+1$. Within each $k$-ring neighborhood, it can be further divided into a series of $(k,t)$-ring neighborhoods with a total length of $T$. {\zlcikm These $(k,t)$-ring sets will be fed into the TRGT module for model training.}


\subsection{TRGT Module}
\label{section:HHGT}
Built upon this innovative $(k,t)$-ring  structure, 
TRGT module seamlessly integrates a Type-level Transformer for aggregating nodes of different types within each $k$-ring neighborhood, followed by a Ring-level Transformer for aggregating different $k$-ring neighborhoods in a hierarchical manner. The diagram of TRGT is shown in Figure~\ref{fig:frame}(b) and the details are elaborated below.

\subsubsection{Type-level Transformer} 
Recall that nodes in HINs come in various types, each representing  distinct concept. Nevertheless, existing HGT-based methodologies tend to mix nodes of different types by {\zlcikm packing all node types into a single sequence and uniformly employing attention over them during neighbor aggregation, which fails to model the distinct roles of nodes with various types, as discussed before.} 

{\zlcikm To overcome this limitation, we design a Type-level Transformer to aggregate neighbors by explicitly considering node type difference. Particularly, given a node $u$ and its neighbors, we first adopt the Ring2Token module to divide the neighbors of node $u$ into several subsets, i.e., $(k,t)$-ring sets. Within each $k$-ring structure, a series of $(k,t)$-ring neighborhoods with a total length of $T$ are extracted. For instance, in Figure~\ref{fig:embedding}(c), three $(k,t)$-ring sets are formed within $P_1$'s 1-ring neighborhood. Here, nodes in the same $(k,t)$-ring are associated with the same node type and semantic function. Therefore, for each $(k,t)$-ring set, the features of all nodes within it are firstly aggregated using an average pooling function to create an embedding token with a specific size $d$, i.e., $x_{k,t} \in \mathbb{R}^{d}$, which explicitly summarizes the information of all nodes with type $t$ within the $k$-ring. In case of an empty set, the embedding token is filled with zeros, ensuring a consistent size of $d$. Thus, each $k$-ring neighborhood can be represented as a sequence of tokens denoted as $x_{k} = \{x_{k,1}, \dots, x_{k,T}\} \in \mathbb{R}^{T\times d}$. Then, we aggregate the representations of different subsets by adopting a Type-level Transformer encoder over the sequence $x_{k}$, as discussed in Section~\ref{sec:encoder}. Through this type-aware aggregation approach, our model can explicitly distinguish neighbors with different types and avoid potential semantic confusion. Note that for the $0$-ring feature $x_0$, we employ a Multi-Layer Perceptron (MLP) to convert the embedding dimension from $\mathbb{R}^{T\times d}$ to $\mathbb{R}^{1\times d}$. 

By stacking $L$ Transformer blocks, we derive the final representation for each $k$-ring neighborhood using a read-out function, denoted as $\{h_0, h_1, \dots, h_K\}$, where $h_0 \in \mathbb{R}^{1\times d}$ and $h_k = \{h_{k,1}, \dots, h_{k,T}\} \in \mathbb{R}^{T\times d}$.}


\noindent \textbf{Type-level Attention Mechanism.} Here, we further introduce a type-level attention mechanism to {\zlcikm serve as the read-out function within each $k$-ring structure.}
Attention function can be described as a mapping between a query and a set of key-value pairs, yielding an output. Specifically, the type-level attention function within each $k$-ring can be defined as follows:
{\qycikm
{\zlcikm
\begin{equation}
\label{eq:att1}
\alpha_t = \frac{exp\left(h_0\cdot h_{k,t}\right)}{\sum_{i=1}^Texp\left(h_0\cdot h_{k,i}\right)}.
\end{equation}
}
Here, $h_0\in \mathbb{R}^{1\times d}$ denotes the $0$-ring representation and $h_{k,t}\in \mathbb{R}^{1\times d}$ denotes the $(k,t)$-ring representation after Transformer encoder. $\alpha_t \in \mathbb{R}^{1}$ is an attention score, $T$ denotes the number of node types and $\cdot$ denotes the dot product.} The final representation of each $k$-ring neighborhood is calculated as:
\begin{equation}
\label{eq:att_read_t}
h_k=\sum_{t=1}^T \alpha_t \cdot h_{kt}.
\end{equation}
Finally, Type-level Transformer outputs a sequence of $k$-ring representations for each node, which later are forwarded into the Ring-level Transformer for representation learning.

\subsubsection{Ring-level Transformer}
Each $k$-ring neighborhood contributes a new layer of information, and their combination provides diverse perspectives, essential for a comprehensive understanding of the HIN. Therefore, the effective collection of information from these $k$-ring neighborhoods is crucial. To tackle this challenge, we develop the Ring-level Transformer for global aggregation across different $k$-ring neighborhoods.
For every node, given a sequence of $k$-ring tokens $\{h_0, h_1, \dots, h_K\}$ obtained from Type-level Transformer, {\zlcikm with each summarizing the unique information of neighbors belonging to a specific $k$-ring structure,} Ring-level Transformer first leverages Transformer encoder to learn the node representations. After the stacking of $L$ Transformer layers, we obtain the representation of each node, i.e., a sequence of representations $\{z_0, z_1, \dots, z_K\}$ where $z_k\in \mathbb{R}^{d}$. 

\noindent
\textbf{Ring-level Attention Mechanism.} 
Considering the potential diverse and unique impacts of different $k$-ring neighborhoods, we design the ring-level attention mechanism for information aggregation. Specifically, it calculates attention coefficients by assessing the relations between the $0$-ring neighborhood (the node itself) and all other $k$-ring neighborhoods, which is formulated as:
{\zlcikm
\begin{equation}
\label{eq:att_r}
\alpha_k=\frac{exp\left((z_0|| z_k)W^T\right)}{\sum_{i=1}^K exp\left((z_0|| z_i)W^T\right)},
\end{equation}
}

where $W \in \mathbb{R}^{1\times 2d}$ denotes the learnable projection, and {\zlcikm $||$} indicates the concatenation operator. Once the attention scores are obtained, they are employed to calculate a linear combination of the corresponding representations, which is written as:
\begin{equation}
\label{eq:att_read_r}
z=z_0+\sum_{k=1}^K{\alpha_k} \cdot {z_k},
\end{equation}
where $K$ represents the number of rings, $z_0$ and $z_k$ denote the representations of $0$-ring neighborhood and $k$-ring neighborhood, respectively. 
Then, we can derive the final representation of each node as $z \in \mathbb{R}^{d}$.

\subsection{Objective Function}
After obtaining the final representations of all nodes through our HHGT model, {\zlcikm we employ the cross-entropy loss to optimize node embeddings in HINs following existing work~\cite{wang2019heterogeneous}. Specifically,} we utilize a classification head to predict the labels of nodes.
This prediction results in a predicted label matrix for nodes, denoted by $\hat{Y} \in \mathbb{R}^{n\times |\mathcal{L}|}$, where $|\mathcal{L}|$ denotes the number of classes. The cross-entropy loss employed is then described as follows:
\begin{equation}
\mathcal{L} = - \sum_{i\in \mathcal{I}}\sum_{j\in \mathcal{L}} Y_{i,j}ln\hat{Y}_{i,j}.
\end{equation}
Here, $\mathcal{I}$ denotes the labeled node set and $Y_{i,j}$ represents the true label. Using the labeled data, we can refine the model through back-propagation, iteratively adjusting parameters to learn the node representations in HINs.


\section{Experiments}
\label{section:experiment}

\begin{table*}[th]
\caption{Dataset Statistics}
\label{tab:data}
\centering
\begin{tabular}{c|ccccccc}
\toprule
Dataset & Objects ($\#$) & $\#$Object & Relations & $\#$Relation & $\#$Label Type & $\#$Labeled Object \\ \hline
ACM & $P$ (4025), $A$ (7167), $S$ (60) & 11,252 & $P\rightleftharpoons A, P\rightleftharpoons S$ & 17,432 & 3 & 4,025 \\
MAG & $P$ (4017), $A$ (15383), $I$ (1480), $F$ (5454) & 26,334 & $P\rightleftharpoons P, P\rightleftharpoons F, P\rightleftharpoons A, A\rightleftharpoons I$ & 86,230 & 4 & 4,017 \\
\bottomrule
\end{tabular}
\end{table*}

\begin{table}[h]
  \caption{Overall evaluation for node classification. The tabular results are in percent; the best results are highlighted in bold; the \underline{underlined} results indicate the second-best performance.}
  \label{tab:nc}
  \vspace{-2mm}
  \setlength\tabcolsep{4pt}
  \centering
  \begin{tabular}{ccccc}
  \toprule
  \multirow{2}{*}{Methods}  & \multicolumn{2}{c}{ACM} & \multicolumn{2}{c}{MAG}  \\
  & Macro-F1 & Micro-F1 & Macro-F1 & Micro-F1 \\ \hline
  PTE & 25.40 & 61.57 & 59.75  & 60.31 \\
  ComplEx & 56.67 & 78.93 & 91.65 & 91.69  \\
  HIN2Vec & 25.40 & 61.57 & 19.21 & 28.08  \\
  M2V & 60.31 & 74.24 & 91.16  & 91.22  \\
  AspEm & 58.72 & 80.70 & 16.85 & 26.11 \\ \hline
  R-GCN & 51.34 & 72.80 & 96.40 & 96.39  \\
  HAN & 60.34 & 81.42 & 97.39 & 97.36  \\ 
  AGAT & 58.14 & 80.75 & 97.50 & 97.51   \\ 
  SHGP & 61.91 & 81.69 & 97.19 & 97.04  \\ \hline
  GTN & 64.49 & 78.26  & 96.63 & 96.64 \\ 
  HGT & \underline{65.48} & 74.91 & 98.80 & 98.76  \\ 
  FastGTN & 64.70 & 76.77 & 93.08 & 93.03  \\ 
  HINormer & 60.31 & \underline{82.98} & \underline{98.85} & \underline{98.88}  \\ \hline
  NAGphormer & 58.94 & 81.49 & 97.76 & 97.76  \\ 
  \hline
  HHGT & \textbf{68.56} & \textbf{83.11} & \textbf{99.25} & \textbf{99.25} \\
  \bottomrule
  \end{tabular}
\end{table}

\begin{table}[h]
  \caption{Overall evaluation for node clustering. The tabular results are in percent; the best results are highlighted in bold; the \underline{underlined} results indicate the second-best performance.}
  \vspace{-2mm}
  \label{tab:nclu}
  \setlength\tabcolsep{10pt}
  \centering
  \begin{tabular}{ccccc}
  \toprule
  
 \multirow{2}{*}{Methods} & \multicolumn{2}{c}{ACM} & \multicolumn{2}{c}{MAG}  \\
 & NMI & ARI & NMI & ARI \\ \hline
  PTE & 0.26 & 0.03 & 20.44 & 10.71  \\
  ComplEx & 29.83 & 25.25 & 73.45 & 71.71 \\
  HIN2Vec & 0.40 & 0.15 & 0.40 & 0.01 \\
  M2V & 39.54 & 32.29 & 3.68 & 2.12 \\
  AspEm  & 39.54 & 32.29 & 0.46 & 0.01 \\ \hline
  R-GCN  & 24.74 & 18.27 & 84.82 & 87.34 \\
  HAN & 50.12 & 50.37 & 84.90 & 88.08 \\ 
  AGAT  & 40.33 & 39.42  & 86.73 & 81.40  \\ 
  SHGP & 39.08 & 32.30 & 86.71 & 88.77 \\ \hline
  GTN & \underline{65.71} & \underline{68.69} & 91.24 & 93.99  \\ 
  HGT & 37.96 & 32.76 & \underline{96.84} & \underline{98.07} \\ 
  FastGTN & 65.68 & 68.65 & 75.62 & 77.31 \\ 
  HINormer & 41.66 & 35.06 & 96.35 & 97.76 \\ \hline
  NAGphormer & 47.30 & 40.05 & 96.22 & 97.61  \\ 
  \hline
  HHGT & \textbf{81.97} & \textbf{88.78} & \textbf{98.65} & \textbf{99.25} \\
  \bottomrule
  \end{tabular}
\end{table}

In this section, we conduct extensive experiments to answer the following research questions: 
\begin{itemize}[leftmargin=*]
    \item \textbf{RQ1:} Can HHGT outperform all baselines across various downstream tasks?
    \item \textbf{RQ2:} What do the learned node embeddings represent? Can these embeddings capture the intricate structures and heterogeneity within HINs?
    \item \textbf{RQ3:} How do different modules of HHGT contribute to enhancing the model performance?
    \item \textbf{RQ4:} How do varying hyper-parameters impact the performance of HHGT?
\end{itemize}

{\qycikm \subsection{Experimental Settings}
\subsubsection{Datasets.} 
We conduct experiments on two publicly available real-world HIN benchmark datasets (i.e. ACM\footnote{\url{https://dl.acm.org/}} and MAG\footnote{\url{https://www.microsoft.com/en-us/research/project/microsoft-academic-graph/}}), which are widely employed in related works \cite{wang2019heterogeneous,ren2019heterogeneous,wang2021self,yang2021interpretable}. The main statistics of datasets are summarized in Table~\ref{tab:data}, and the details are shown below.

\begin{itemize}[leftmargin=*]
\item \textbf{ACM} is a subgraph of ACM digital library, which contains 4,025 papers (P), 7,167 authors (A) and 60 subjects (S). All objects have 128-dimensional features. There are two link types, i.e., 13,407 Paper-Author (PA) {\qycikm links} and 4,025 Paper-Subject (PS) {\qycikm links} between all kinds of objects. Papers (P), the target nodes, are {\qycikm classified into three categories}: Computer Network, Data Mining and Database, according to their fields.
\item \textbf{MAG} is extracted from Microsoft Academic Graph, which contains 4,017 papers (P), 15,383 authors (A), 1,480 institutions (I) and 5,454 fields (F). All objects have attributes with 128-dimensional features. It also contains four relation types: 3,880 Paper-Paper (PP) {\qycikm links}, 40,378 Paper-Field (PF) {\qycikm links}, 26,144 Paper-Author (PA) {\qycikm links} and 15,468 Author-Institution (AI) {\qycikm links}. {\qycikm The target nodes, which are the papers (P),} are divided into four groups based on their published venues: Astrophysics, IEEE Journal of Photovoltaics, Journal of Applied Meteorology and Climatology, and Low Temperature Physics.
\end{itemize}

\subsubsection{Baselines.} 

To verify the effectiveness of our model, we compare our HHGT with two groups of baselines: shallow model-based methods and deep model-based methods. The former group includes PTE~\cite{tang2015pte}, ComplEx~\cite{trouillon2016complex}, HIN2Vec~\cite{fu2017hin2vec}, M2V~\cite{dong2017metapath2vec}, AspEm~\cite{shi2018aspem}. The latter group can be further divided into $(1)$ HGNN-based models including R-GCN~\cite{schlichtkrull2018modeling}, HAN~\cite{wang2019heterogeneous}, AGAT~\cite{liu2022aspect}, SHGP~\cite{yang2022self}; $(2)$ GT-based models including one homogeneous GT-based method NAGphormer~\cite{chen2022nagphormer} and three heterogeneous GT-based methods {\zqycikm including} HGT~\cite{hu2020heterogeneous}, FastGTN~\cite{yun2022graph} and HINormer~\cite{mao2023hinormer}. 
{\zqycikm Here, SHGP, PTE, ComplEx, HIN2Vec, M2V and AspEm are unsupervised methods, while R-GCN, HAN, AGAT, GTN, HGT, FastGTN, HINormer and NAGphormer are semi-supervised methods, {\zlcikm which are the same with our model.}}


\subsubsection{Reproducibility.} For the proposed HHGT, we optimize the model with Adam. We set the dropout rate, attention dropout rate, weight decay and head number as 0.01, 0.05, 0.00 and 8 for both datasets, respectively. The learning rate is searched from 1e-4 to 1e-2. For all compared baselines, we employ their publicly released source code 
and adopt the hyper-parameters recommended in their papers to ensure consistency. For all methods, we set the hidden dimension as 128 for ACM dataset and 512 for MAG dataset for a fairness comparison. 
{\zqycikm To ensure reproducibility, we include our source code, datasets, as well as the instructions to the selected baselines, in an anonymous repository~\footnote{\url{https://anonymous.4open.science/r/HHGT-D78B}}.}
All the experiments are conducted on a  Linux (Ubuntu 18.04.6 LTS) server with one GPU (NVIDIA Tesla V100-SXM2) and two CPUs (Intel Xeon E5-2698 v4). 


{\zlcikm \smallskip\noindent\textbf{Remarks.}} Following previous work~\cite{wang2019heterogeneous}, we employ the cross-entropy loss to optimize node embeddings in HINs. Once our HHGT is trained, we can get all node embeddings via feed forward.
Consistent with the settings of prior works~\cite{wang2019heterogeneous, yang2022self, zhao2020network}, we employ node classification and node clustering as downstream tasks to evaluate the quality of the learned node embeddings. This setting allows the learned node embeddings to serve as as a universal feature representation, which enables our model to be applied to various tasks without the need for retraining. }

\subsection{Node Classification (RQ1)}
\textbf{Settings.} Node classification task aims to assign categories to nodes within a network. Following \cite{hu2020heterogeneous}, we train a separate linear Support Vector Machine (LinearSVC) \cite{fan2008liblinear} with 80$\%$ of the labeled nodes and predict on the remaining 20$\%$ data. We repeat the process 10 times and report the average results. Micro-F1 and Macro-F1 scores are adopted to evaluate the effectiveness. 

\noindent \textbf{Results.} The overall experimental results are shown in Table~\ref{tab:nc}. 
As observed, HHGT achieves the best overall performance, which indicates its superior effectiveness. The main reasons for the observed improvement may include: (1) The Type-level Transformer optimally utilizes node type information during neighbor aggregation, enabling the effective capture of proper correlations between nodes of different types; (2) The Ring-level Transformer considers the distinctions between neighbors at different distances, facilitating powerful neighbor aggregation across various hierarchical levels. 
{\zlcikm Additionally, we observe that the second best is relatively unstable, demonstrating that our model is very robust against various settings as well as metrics. Among the baselines, we can observe that deep model-based approaches generally outperform their shallow model-based counterparts, highlighting the benefits of multi-layered feature extraction in capturing complex information within HINs.}


\begin{figure*}[thb]    
  \centering
  \subfloat[HHGT]{\label{fig:acm_HHGT}\parbox{0.125\textwidth}{\includegraphics[width=0.07\textwidth]{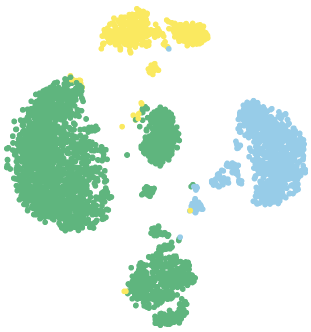}}}
  \subfloat[NAGphormer]{\label{fig:acm_NAG}\parbox{0.125\textwidth}{\includegraphics[width=0.07\textwidth]{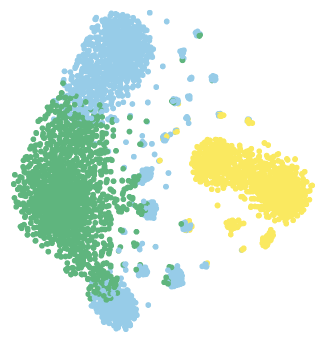}}}
  \subfloat[HINormer]{\label{fig:acm_HINormer}\parbox{0.125\textwidth}{\includegraphics[width=0.07\textwidth]{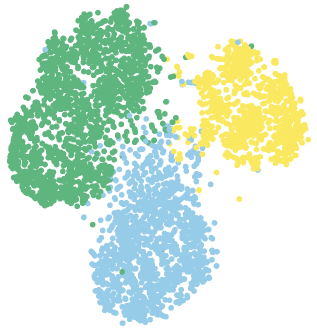}}}
  \subfloat[FastGTN]{\label{fig:acm_FAST}\parbox{0.125\textwidth}{\includegraphics[width=0.07\textwidth]{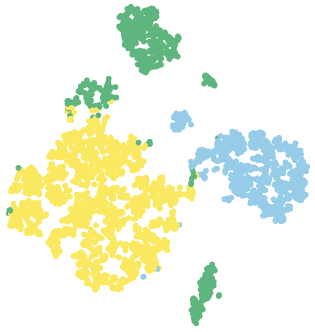}}}
  \subfloat[HGT]{\label{fig:acm_HGT}\parbox{0.125\textwidth}{\includegraphics[width=0.07\textwidth]{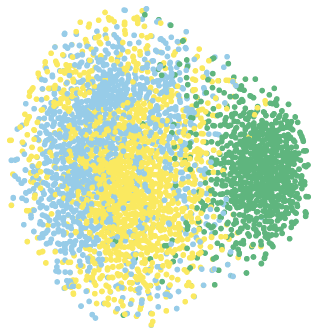}}}
  \subfloat[GTN]{\label{fig:acm_GTN}\parbox{0.125\textwidth}{\includegraphics[width=0.07\textwidth]{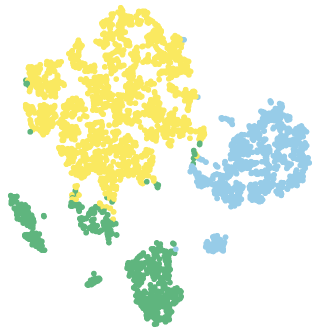}}}
  \subfloat[SHGP]{\label{fig:acm_SHGP}\parbox{0.125\textwidth}{\includegraphics[width=0.07\textwidth]{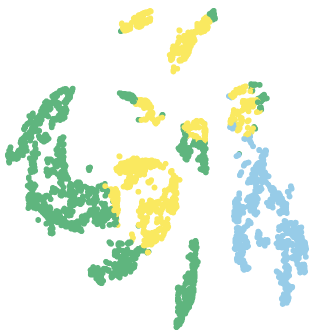}}}
  \subfloat[AGAT]{\label{fig:acm_AGAT}\parbox{0.125\textwidth}{\includegraphics[width=0.07\textwidth]{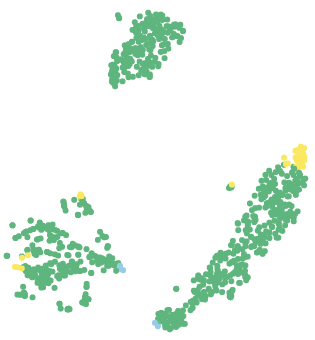}}}

  
  \subfloat[HAN]{\label{fig:acm_HAN}\parbox{0.125\textwidth}{\includegraphics[width=0.07\textwidth]{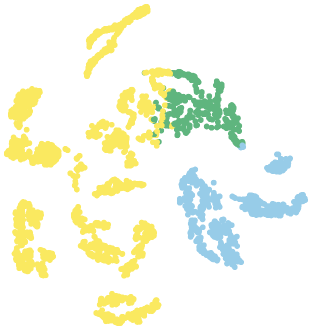}}}
  \subfloat[R-GCN]{\label{fig:acm_RGCN}\parbox{0.125\textwidth}{\includegraphics[width=0.07\textwidth]{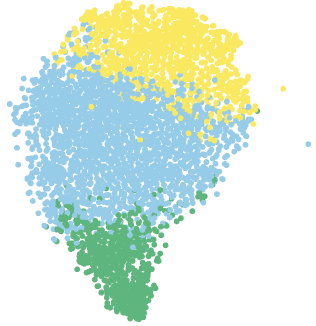}}}
  \subfloat[AspEm]{\label{fig:acm_AspEm}\parbox{0.125\textwidth}{\includegraphics[width=0.07\textwidth]{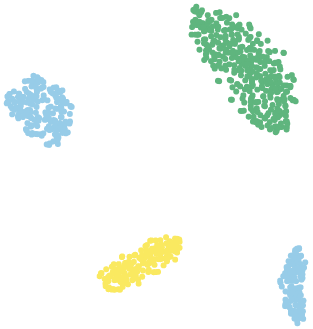}}}
  \subfloat[M2V]{\label{fig:acm_M2V}\parbox{0.125\textwidth}{\includegraphics[width=0.07\textwidth]{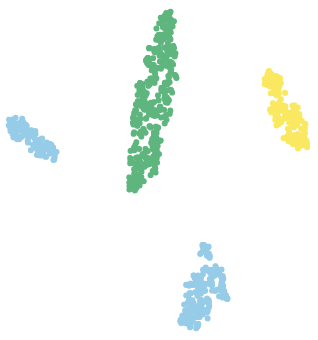}}}
  \subfloat[HIN2Vec]{\label{fig:acm_HIN2Vec}\parbox{0.125\textwidth}{\includegraphics[width=0.07\textwidth]{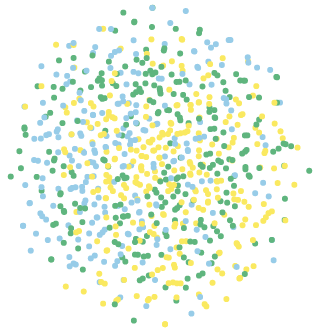}}}
  \subfloat[ComplEx]{\label{fig:acm_ComplEx}\parbox{0.125\textwidth}{\includegraphics[width=0.07\textwidth]{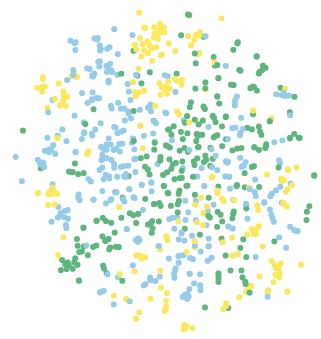}}}
  \subfloat[PTE]{\label{fig:acm_PTE}\parbox{0.125\textwidth}{\includegraphics[width=0.07\textwidth]{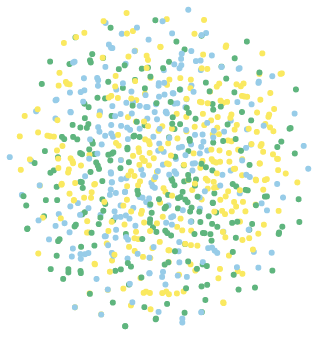}}}

  \caption{{\zqycikm Embedding visualization on ACM dataset, where our HHGT model shows much higher intra-class similarity.}}    
  \label{fig:vis_acm}  
\end{figure*}

\begin{figure*}[thb]    
  \centering
  \subfloat[HHGT]{\label{fig:mag_HHGT}\parbox{0.125\textwidth}{\includegraphics[width=0.07\textwidth]{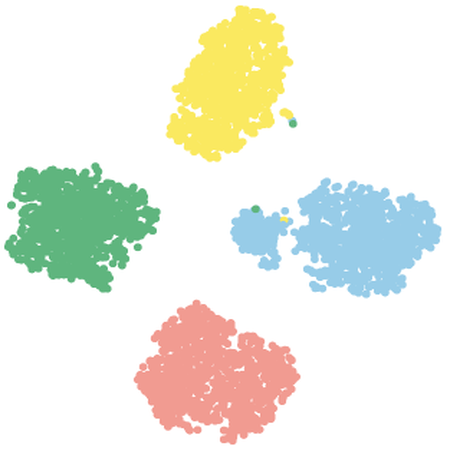}}}
  \subfloat[NAGphormer]{\label{fig:mag_NAG}\parbox{0.125\textwidth}{\includegraphics[width=0.07\textwidth]{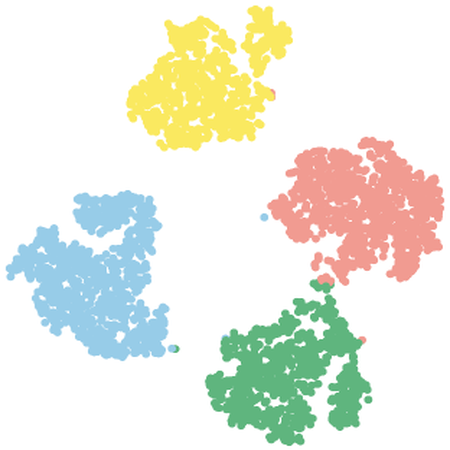}}}
  \subfloat[HINormer]{\label{fig:mag_HINormer}\parbox{0.125\textwidth}{\includegraphics[width=0.07\textwidth]{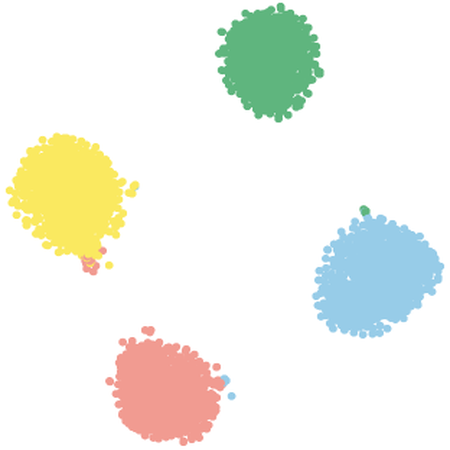}}}
  \subfloat[FastGTN]{\label{fig:mag_FAST}\parbox{0.125\textwidth}{\includegraphics[width=0.07\textwidth]{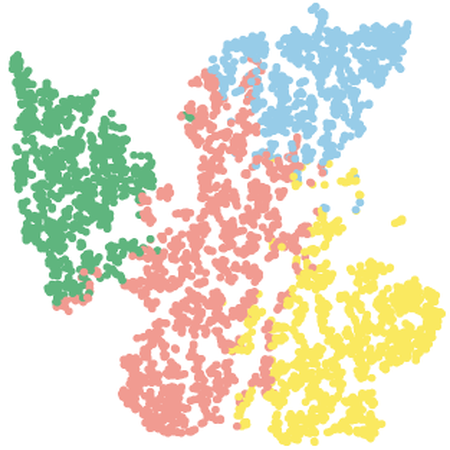}}}
  \subfloat[HGT]{\label{fig:mag_HGT}\parbox{0.125\textwidth}{\includegraphics[width=0.07\textwidth]{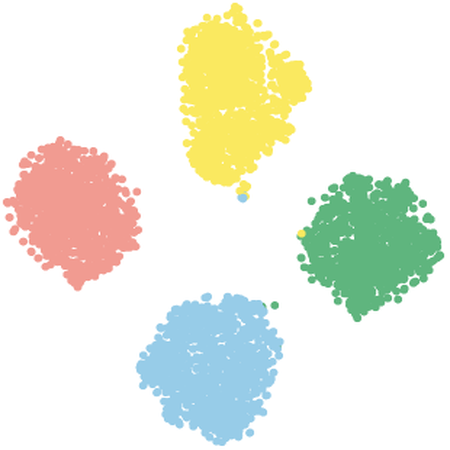}}}
  \subfloat[GTN]{\label{fig:mag_GTN}\parbox{0.125\textwidth}{\includegraphics[width=0.07\textwidth]{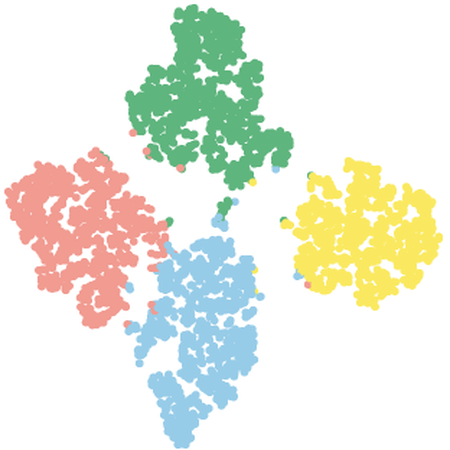}}}
  \subfloat[SHGP]{\label{fig:mag_SHGP}\parbox{0.125\textwidth}{\includegraphics[width=0.07\textwidth]{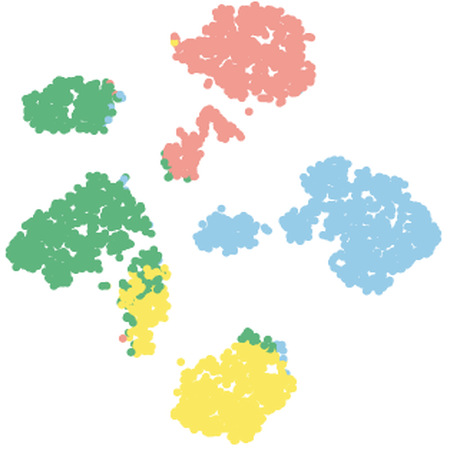}}}
  \subfloat[AGAT]{\label{fig:mag_AGAT}\parbox{0.125\textwidth}{\includegraphics[width=0.07\textwidth]{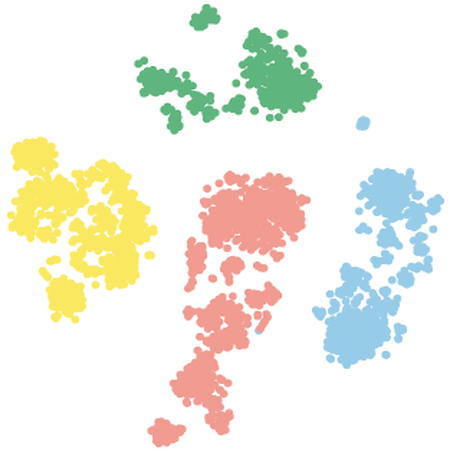}}}

  
  \subfloat[HAN]{\label{fig:mag_HAN}\parbox{0.125\textwidth}{\includegraphics[width=0.07\textwidth]{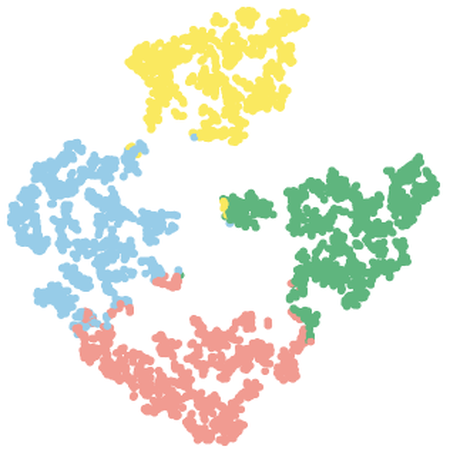}}}
  \subfloat[R-GCN]{\label{fig:mag_RGCN}\parbox{0.125\textwidth}{\includegraphics[width=0.07\textwidth]{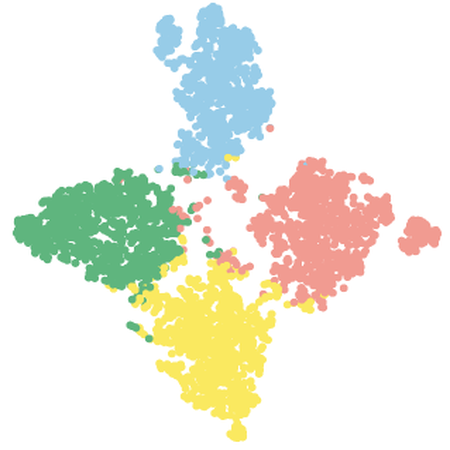}}}
  \subfloat[AspEm]{\label{fig:mag_AspEm}\parbox{0.125\textwidth}{\includegraphics[width=0.07\textwidth]{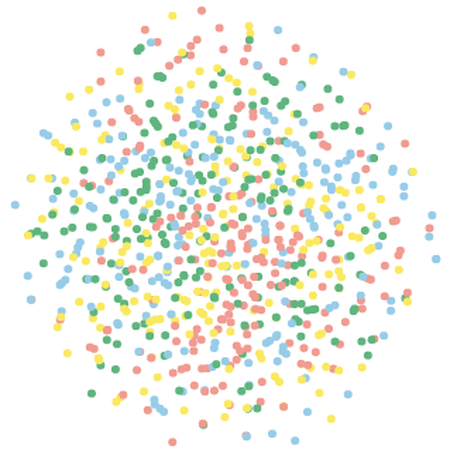}}}
  \subfloat[M2V]{\label{fig:mag_M2V}\parbox{0.125\textwidth}{\includegraphics[width=0.07\textwidth]{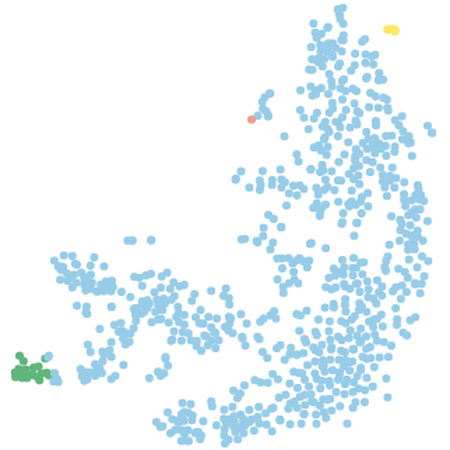}}}
  \subfloat[HIN2Vec]{\label{fig:mag_HIN2Vec}\parbox{0.125\textwidth}{\includegraphics[width=0.07\textwidth]{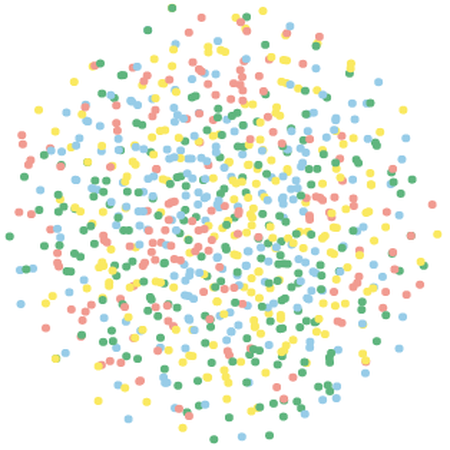}}}
  \subfloat[ComplEx]{\label{fig:mag_ComplEx}\parbox{0.125\textwidth}{\includegraphics[width=0.07\textwidth]{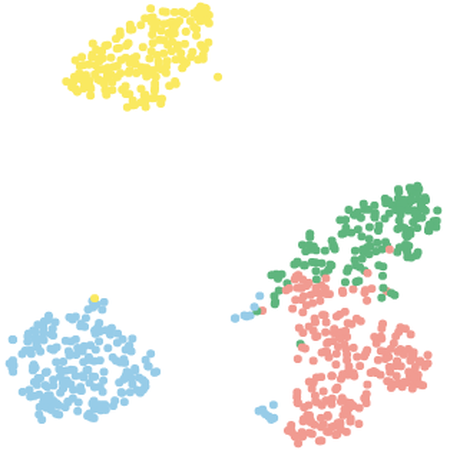}}}
  \subfloat[PTE]{\label{fig:mag_PTE}\parbox{0.125\textwidth}{\includegraphics[width=0.07\textwidth]{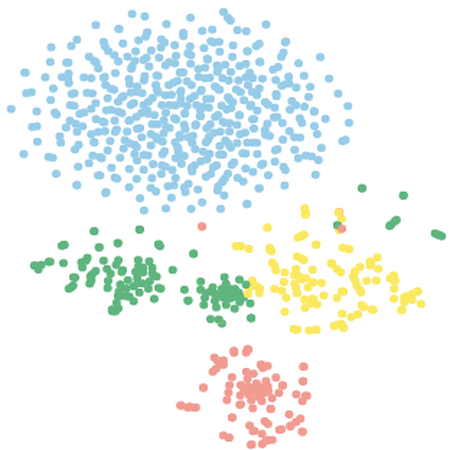}}}

  \caption{Embedding visualization on MAG dataset, {\zqycikm  where our HHGT model clearly separates papers from different published venues with well-defined boundaries.}}    
  \label{fig:vis_mag}  
\end{figure*}

\begin{figure}[h]    
  \centering          
  \subfloat[Micro-F1]  {\label{fig:att_nc_mi}\includegraphics[width=0.2\textwidth]{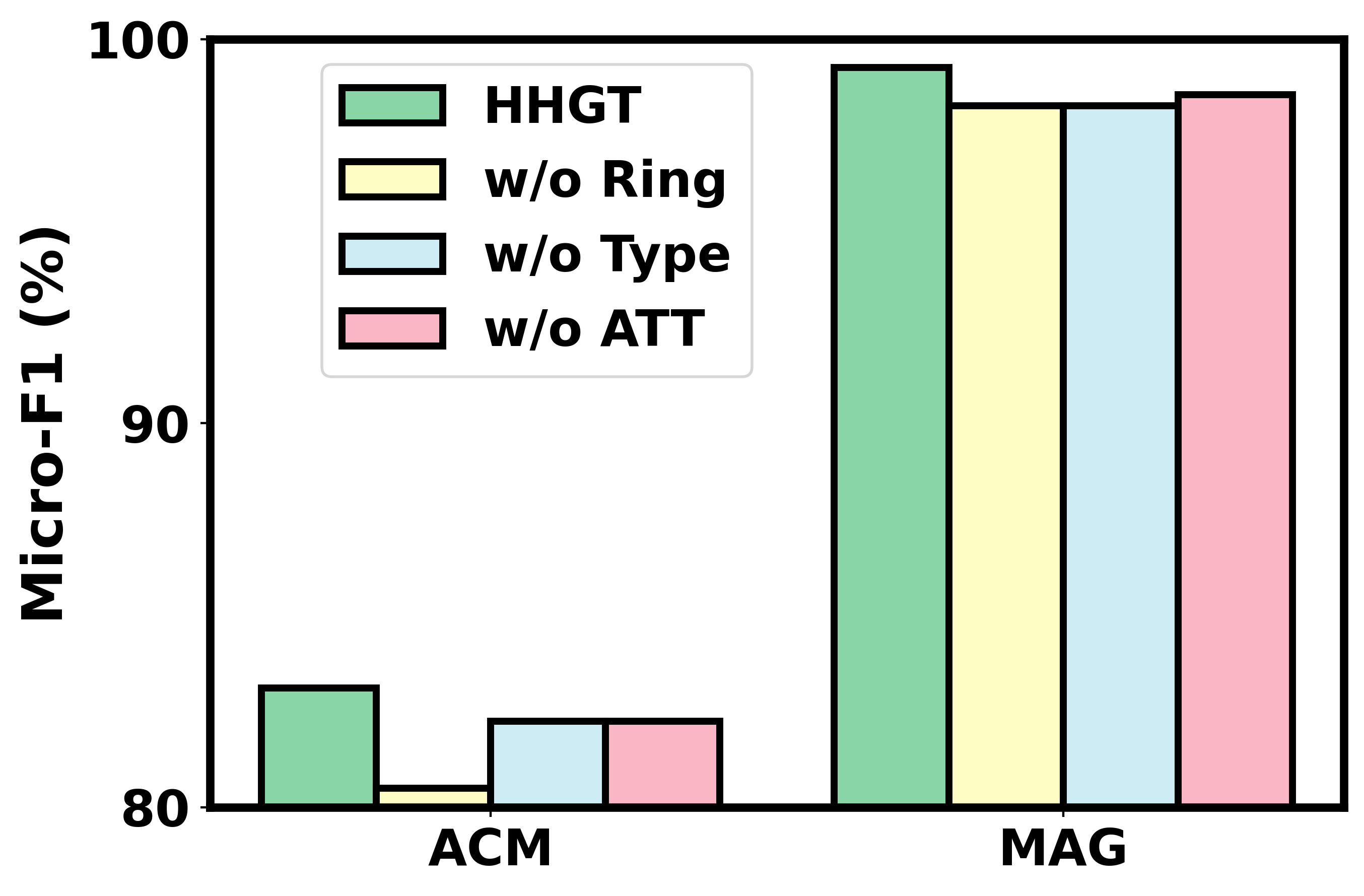}
  }
  \subfloat[Macro-F1]{\label{fig:att_nc_ma}\includegraphics[width=0.2\textwidth]{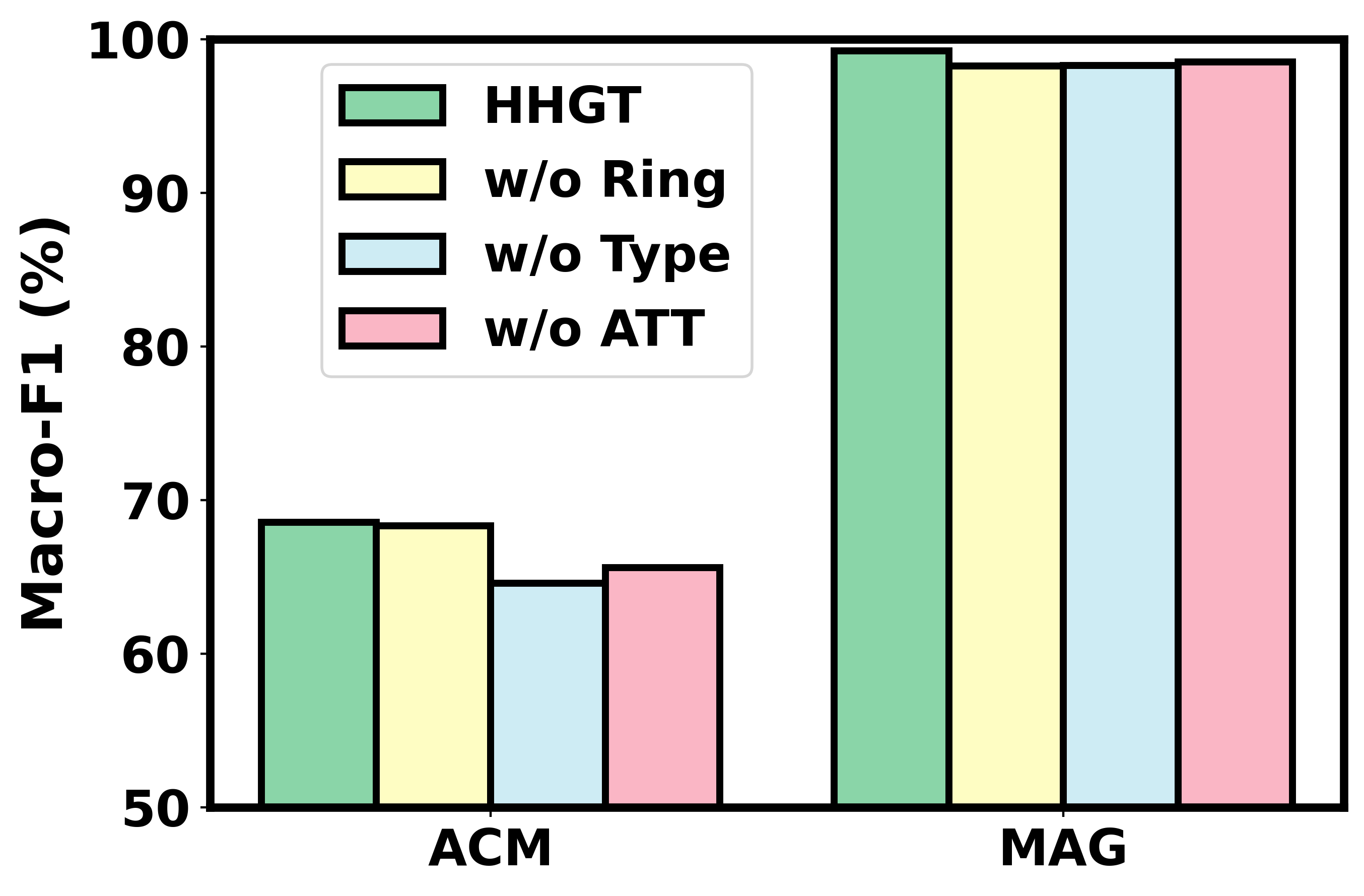}
  }
  \vspace{-3mm}
  \caption{Ablation study for node classification.}    
  \label{fig:nc_att}  
  \vspace{-3mm}
\end{figure}

\begin{figure}[h]    
  \centering          
  \subfloat[NMI]  {\label{fig:att_nc_nmi}\includegraphics[width=0.2\textwidth]{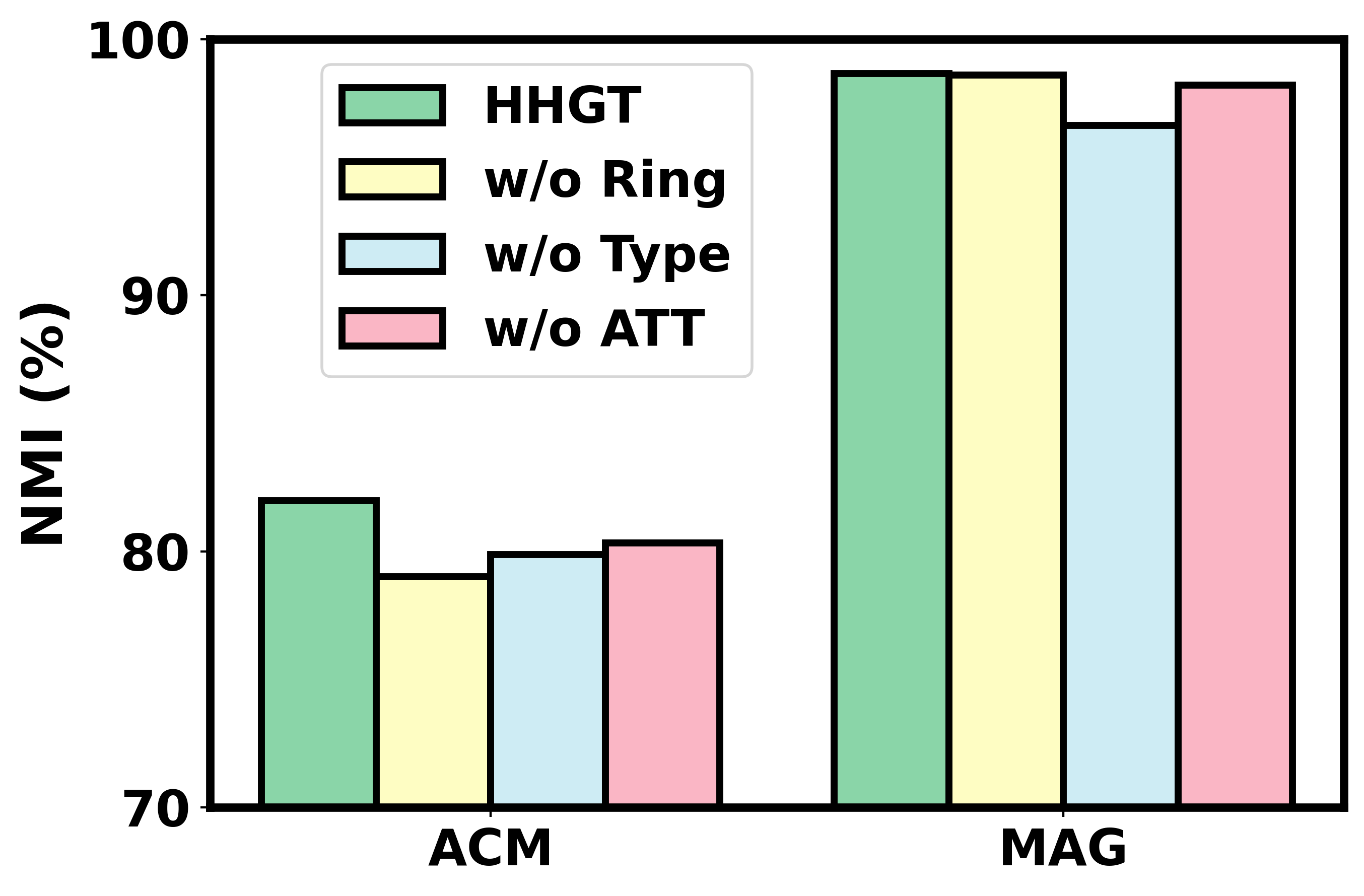}
  }
  \subfloat[ARI]{\label{fig:att_nc_ari}\includegraphics[width=0.2\textwidth]{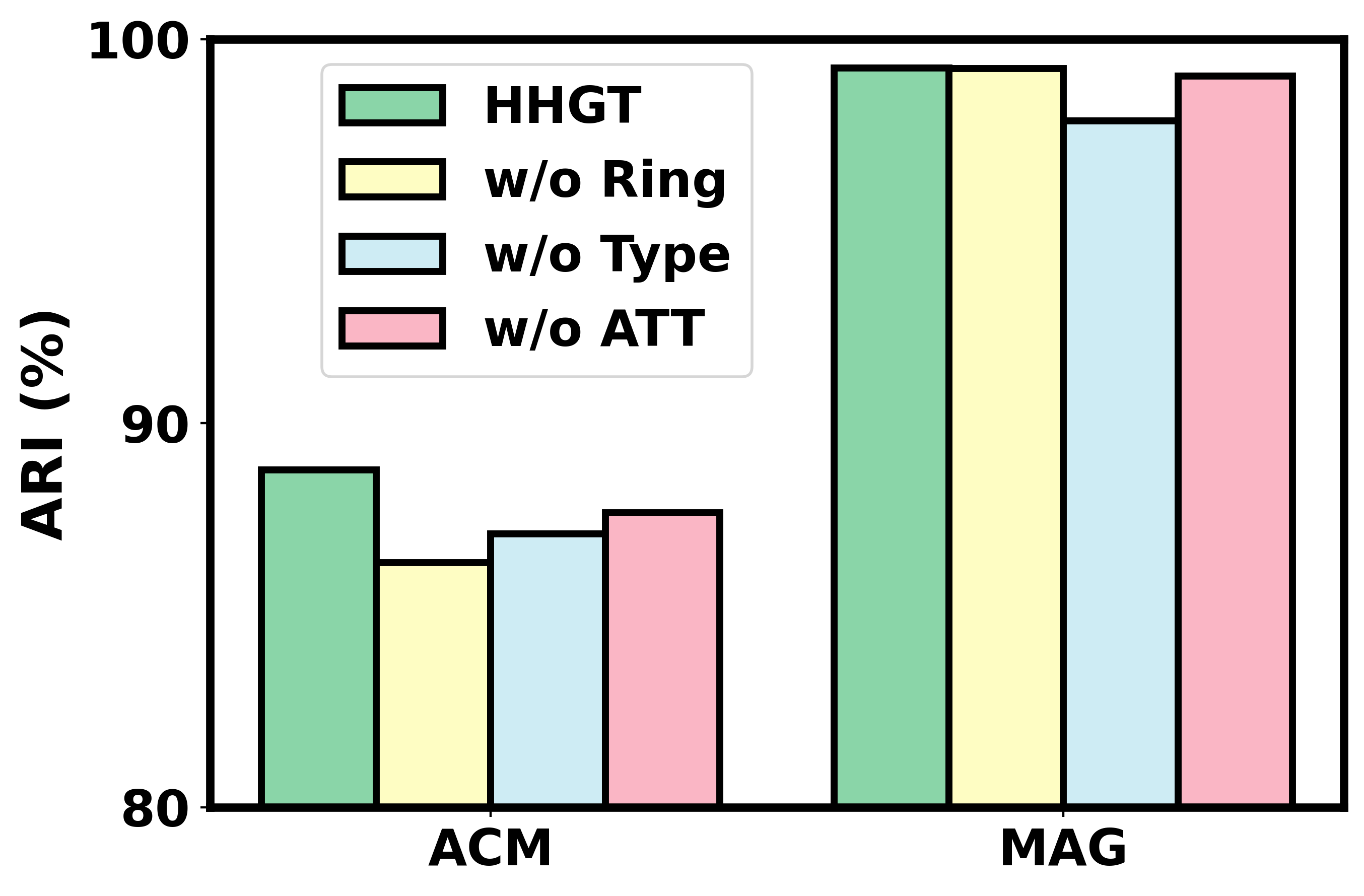}
  }
  \vspace{-3mm}
  \caption{Ablation study for node clustering.} 
  \label{fig:nclu_att}  
  \vspace{-3mm}
\end{figure}

\subsection{Node Clustering (RQ1)}
\textbf{Settings.} Node clustering task seeks to group nodes in a network, according to common structural and attribute features. Following \cite{yang2022self,ren2019heterogeneous,wang2019heterogeneous,zhao2020network}, we adopt an unsupervised learning set where the input is solely node embeddings, and we leverage K-Means to cluster nodes based on their representations generated by the model. The number of clusters for K-Means is set as the category number, and we utilize the same ground truth as employed in node classification task. NMI and ARI are utilized as evaluation metrics here{\qy, following \cite{wang2019heterogeneous,yang2022self}}. Due to the sensitivity of K-Means to initial centroids, we conduct the process 10 times and report the average results.

\noindent \textbf{Results.} Table~\ref{tab:nclu} demonstrates the overall results of all approaches for node clustering task. As we can see, our HHGT consistently surpasses all baseline methods in node clustering across various HINs, demonstrating substantial improvements in both NMI and ARI metrics. 
{\qycikm Specifically, HHGT exhibits an improvement of up to 24.75$\%$ in NMI and 29.25$\%$ in ARI over the top-performing baselines on the ACM dataset, respectively.}
This is because our model simultaneously incorporates Ring-level Transformer and Type-level Transformer, where the former aids in capturing the differences between neighbors at different distances while the latter emphasizes the importance of node types during neighbor aggregation. The hierarchical integration of these two Transformers enables the model to synthesize information at multiple levels, enhancing the diversity and richness of node representations. {\zlcikm Among the baselines, we can observe that GT-based models generally perform much better than HGNN-based models, demonstrating the advantages of the graph transformer architecture in HIN representation learning. 
Furthermore, the performance of heterogeneous GT-based methods such as GTN and HGT is better than that of their homogeneous GT-based counterpart NAGphormer in many cases, verifying the importance of considering the relation heterogeneity within HINs. Our model also benefits from this aspect by designing the $(k,t)$-ring neighborhood structure to emphasize the inherent heterogeneity of both distance and types within HINs.}


\subsection{Embedding Visualization (RQ2).} 
\noindent \textbf{Settings.} For a more intuitive comparison, we conduct embedding visualization to represent an HIN in a low-dimensional space. The goal is to learn node embeddings using the HIN representation learning model and project them into a 2-dimensional space. We employ the t-SNE~\cite{fan2008liblinear} technique for visualization, with a specific focus on paper representations
on {\qycikm both datasets. Nodes are color-coded based on fields and published venues for ACM and MAG, respectively.} 

\noindent \textbf{Results.} The results are shown in {\zlcikm Figure \ref{fig:vis_acm} and} Figure~\ref{fig:vis_mag}, from which we can find the following phenomenons: 
{1) \zlcikm The shallow model-based methods always show mixed patterns among papers from various venues, lacking clear clustering boundaries. For example, in HIN2Vec, all papers are mixed together in both datasets. This is due to their limited modeling ability to capture intricate structural and semantic relations within HINs. 2) Among all baselines, HINormer and GTN, two HGT-based models, provide more reasonable visualization results. However, some clusters are still mixed with each other, and there is no clear margin between different classes, especially in the ACM dataset, as shown in Figure~\ref{fig:acm_HINormer} and Figure~\ref{fig:acm_GTN}. 3) In contrast, the visualization of HHGT reveals high intra-class similarity, distinctly separating papers from different published venues with well-defined boundaries. This clustered structure signifies the tight connections and semantic similarities, showcasing the effectiveness of our proposed HHGT model.}


\subsection{Ablation Study (RQ3)} 
{\zlcikm In order to understand the impact of different components within the proposed framework on the overall performances, we conduct ablation studies by removing or replacing key modeling modules of {\zqycikm HHGT} on two datasets. Specifically, we focus on three key modules: {\zqycikm (1)} the $k$-ring neighborhood structure and its corresponding Ring-level Transformer for distance heterogeneity modeling, {\zqycikm (2)} the $(k,t)$-ring neighborhood structure and its corresponding Type-level Transformer for further type heterogeneity modeling, {\zqycikm (3)} and the attention-based readout function upon two Transformer encoders. By removing or replacing these modules, we can obtain different variants of HHGT as follows:

\begin{itemize}[leftmargin=*]
    \item \textbf{w/o Ring:} In this variant, we replace our $k$-ring structures with traditional $k$-hop patterns for neighbor extraction without partitioning them into different distance-based non-overlapping subsets, and the Ring-level Transformer is then utilized upon the extracted hop-based neighborhood structure. 
    \item \textbf{w/o Type:} Within each $k$-ring structure, this variant mixes all neighbors of different node types without further partitioning them into different type-based subsets, and removes the Type-level Transformer module. 
    \item \textbf{w/o ATT:} In this variant, we replace the attention-based readout functions defined in Equation (\ref{eq:att_read_t}) and Equation (\ref{eq:att_read_r}) with average pooling functions. 
\end{itemize}

}


The results for node classification and node clustering on {\zqycikm both} datasets are illustrated in Figure~\ref{fig:nc_att} and Figure~\ref{fig:nclu_att}, respectively. Based on the results, we have the following observations: 
(1) The {\zlcikm model performance} significantly decreases across both datasets when the $k$-ring structure is replaced with the {\zlcikm traditional} $k$-hop pattern (i.e., {\zlcikm HHGT vs. w/o Ring}), {\zlcikm which emphasizes the importance of the proposed $k$-ring structure. The reason is that the $k$-ring structure, integral to HHGT, excels in capturing distance heterogeneity within HINs by effectively differentiating between neighbors at varying distances. (2) HHGT consistently outperforms w/o Type in all metrics on both datasets, which demonstrates the effectiveness of our Type-level Transformer module. The results also highlight the importance of explicitly considering type heterogeneity in HIN representation learning by further partitioning $k$-ring to $(k,t)$-ring structure. (3) w/o ATT shows inferior performance compared to HHGT across two downstream tasks and two datasets, indicating that the proposed attention-based readout function is beneficial for learning more general and expressive node representations.}

\begin{figure}[h]    
  \centering          
  \subfloat[Micro-F1]  {\label{fig:nc_d_mi}\includegraphics[width=0.2\textwidth]{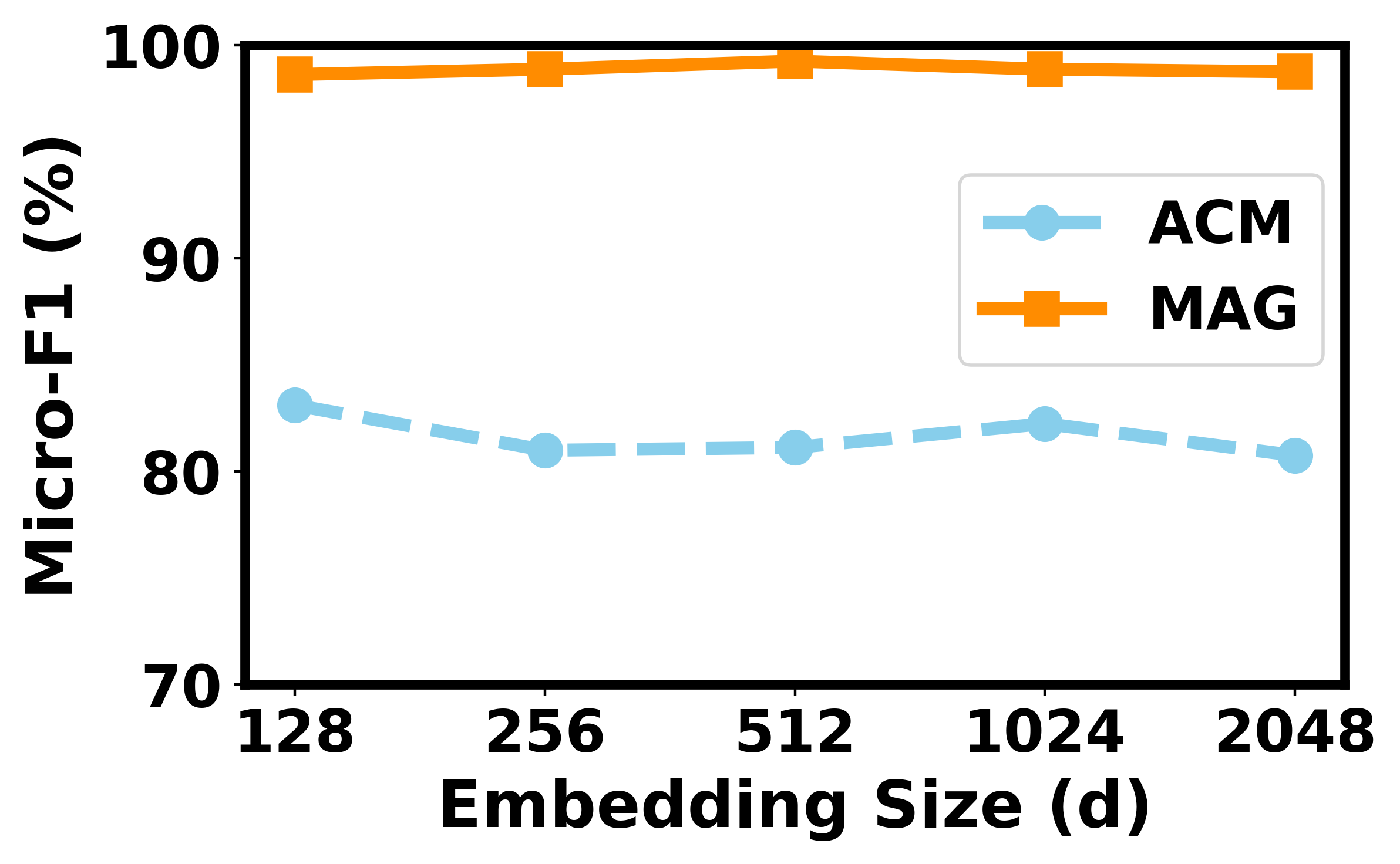}}
  \subfloat[Macro-F1]{\label{fig:nc_d_ma}\includegraphics[width=0.2\textwidth]{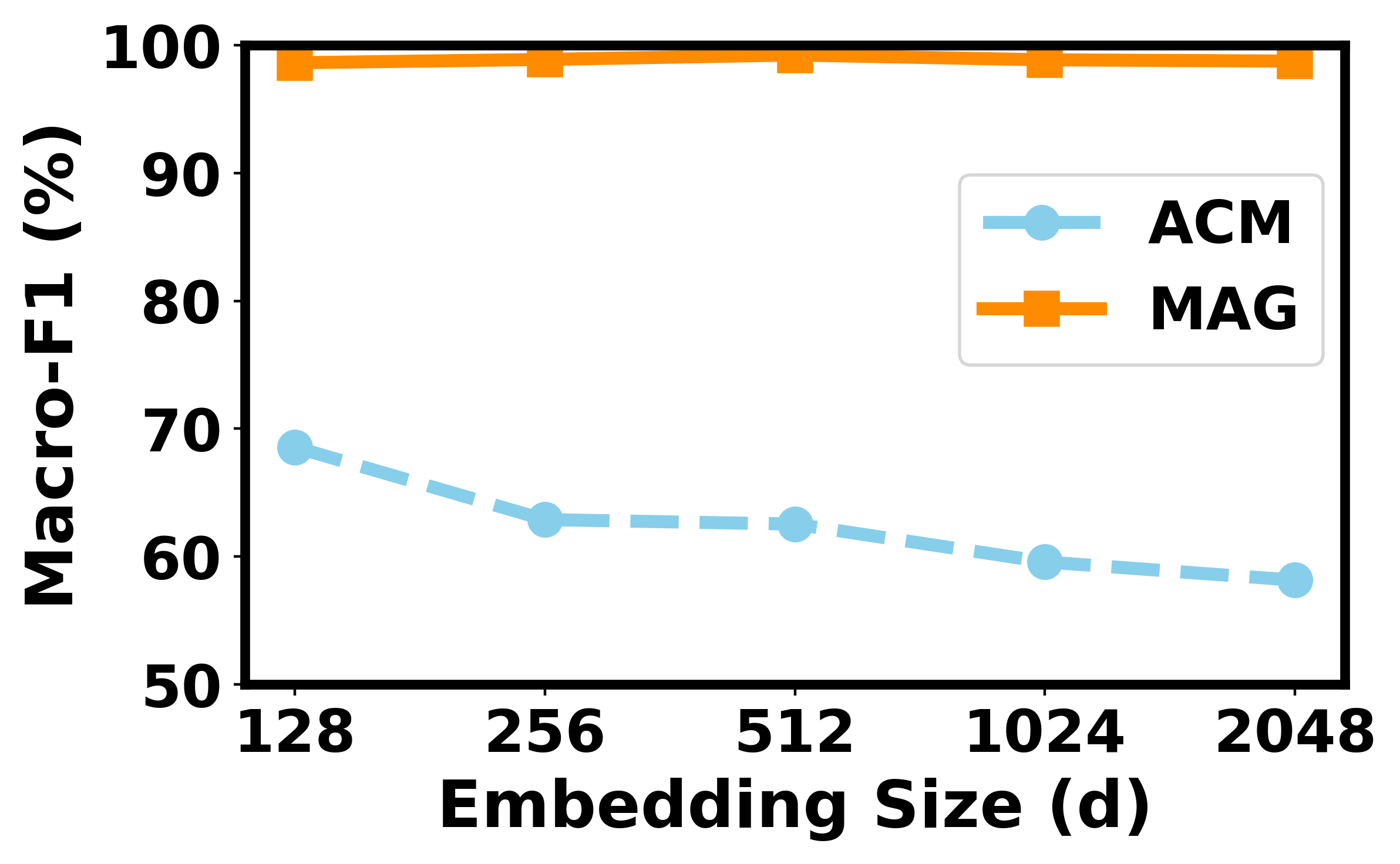}}

  \caption{Embedding size study for node classification.}    
  \label{fig:cla_d}  
\end{figure}
\begin{figure}[h]    
  \centering          
  \subfloat[Micro-F1]  {\label{fig:nc_k_mi}\includegraphics[width=0.2\textwidth]{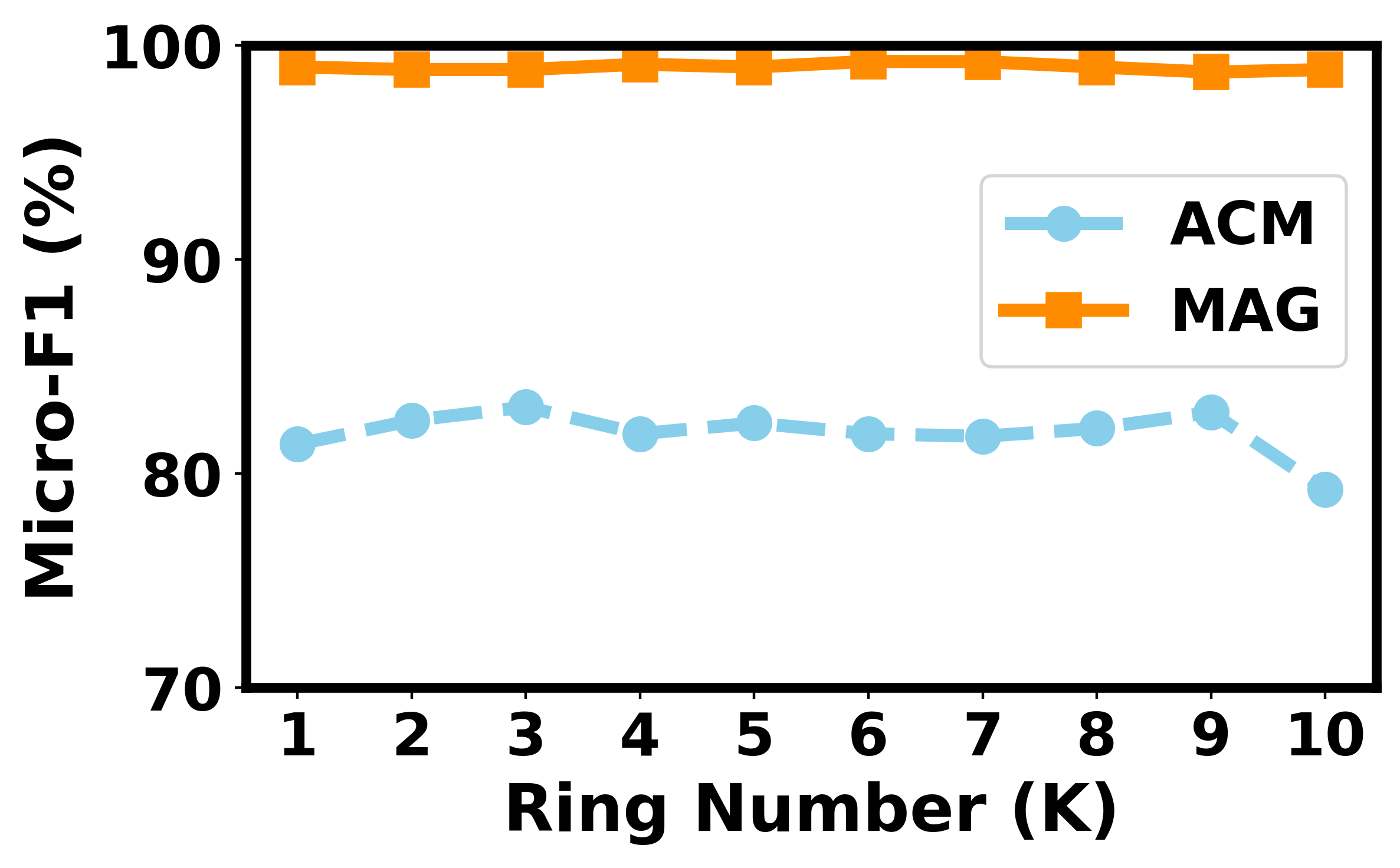}
  }
  \subfloat[Macro-F1]{\label{fig:nc_k_ma}\includegraphics[width=0.2\textwidth]{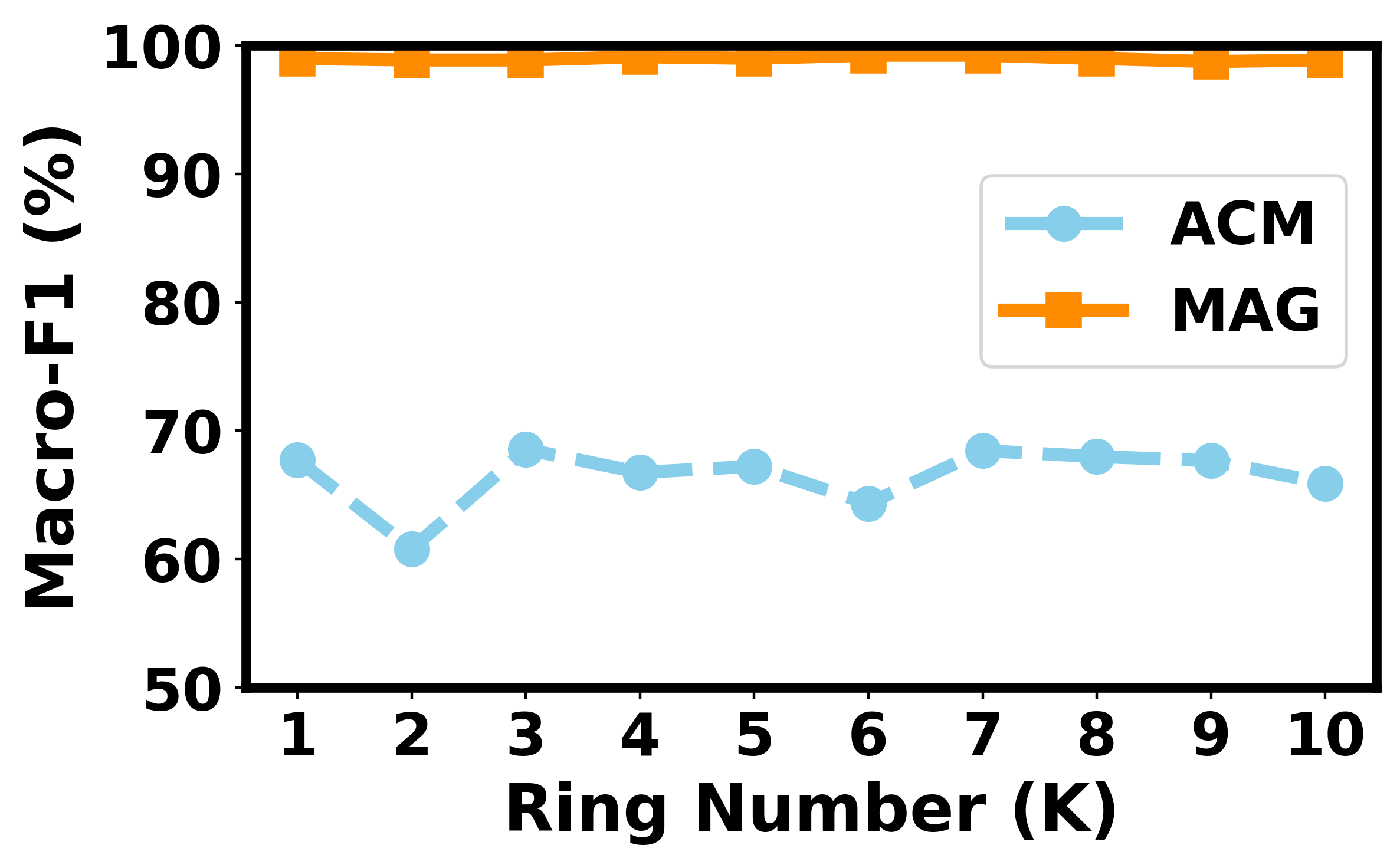}
  }
  \vspace{-3mm}
  \caption{Ring number study for node classification.} 
  \label{fig:cla_K}  
\end{figure}

{\qycikm 
\subsection{Parameter Study (RQ4)} 
We investigate the sensitivity of HHGT with respect to four key hyper-parameters, i.e., the Ring-level Transformer layer number $Lh$, the Type-level Transformer layer number $Lt$, the embedding size $d$ and the number of rings $K$. The results of node classification with different settings on both datasets are depicted in Figures~\ref{fig:cla_lt}-\ref{fig:cla_K}. 

\begin{figure}[h]    
  \centering          
  \subfloat[Micro-F1]  {\label{fig:nc_lh_mi}\includegraphics[width=0.2\textwidth]{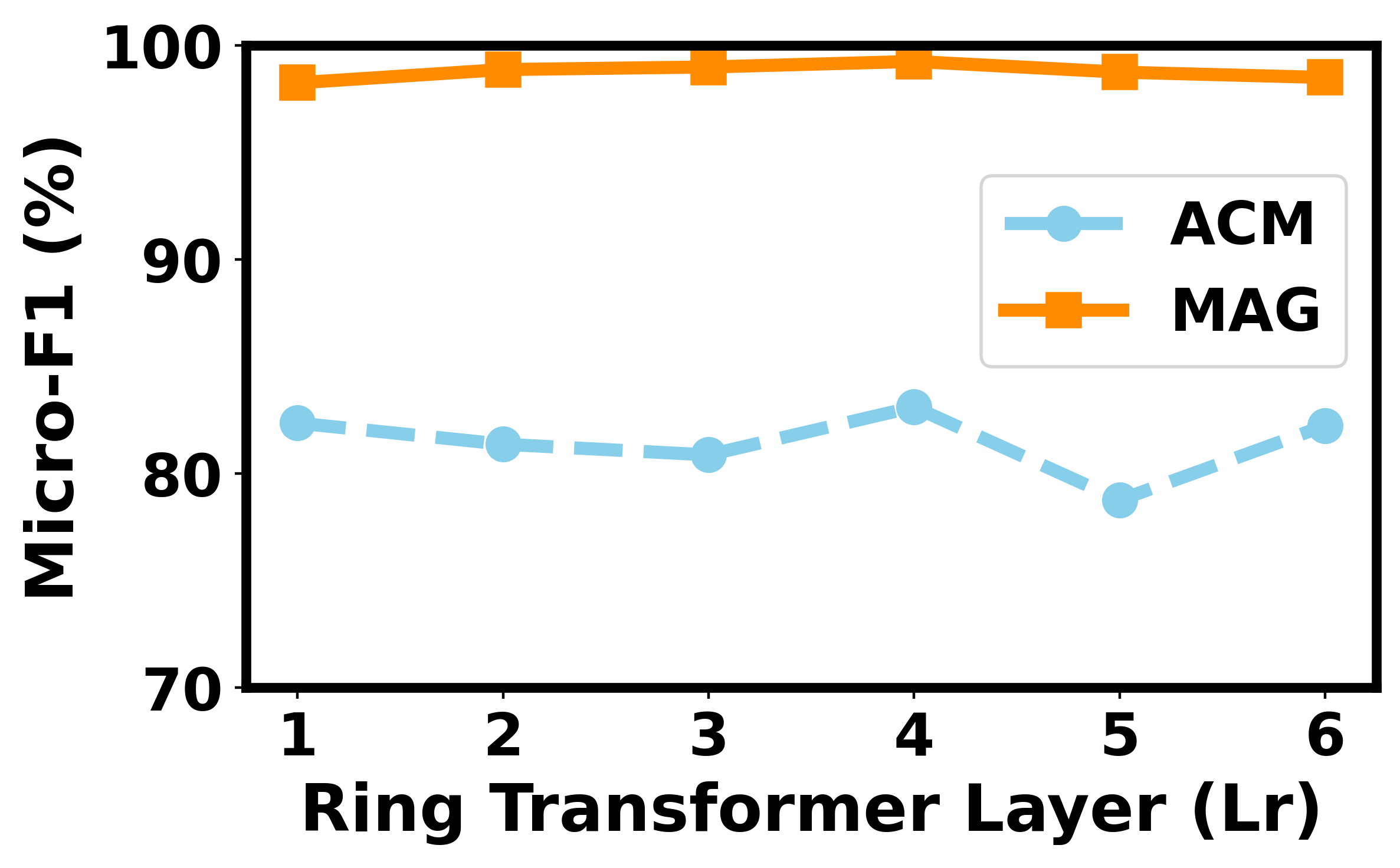}}
  \subfloat[Macro-F1]{\label{fig:nc_lh_ma}\includegraphics[width=0.2\textwidth]{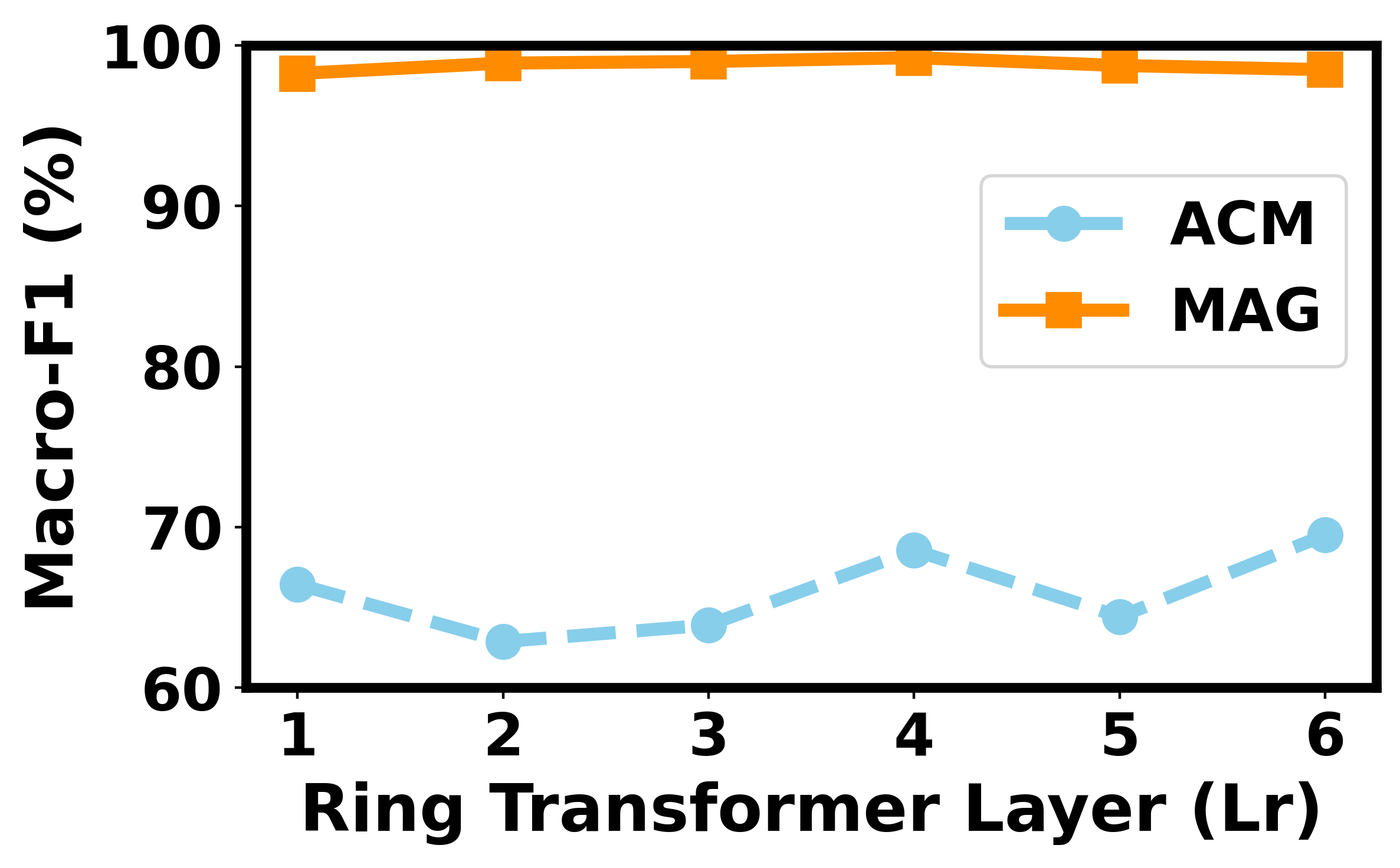}}
  \caption{{\zlcikm Ring-level} transformer layer {\zqycikm number} study for node classification.}   
  \label{fig:cla_lh}  
\end{figure}

\vspace{-4mm}

\begin{figure}[h]    
  \centering          
  \subfloat[Micro-F1]  {\label{fig:nc_lt_mi}\includegraphics[width=0.2\textwidth]{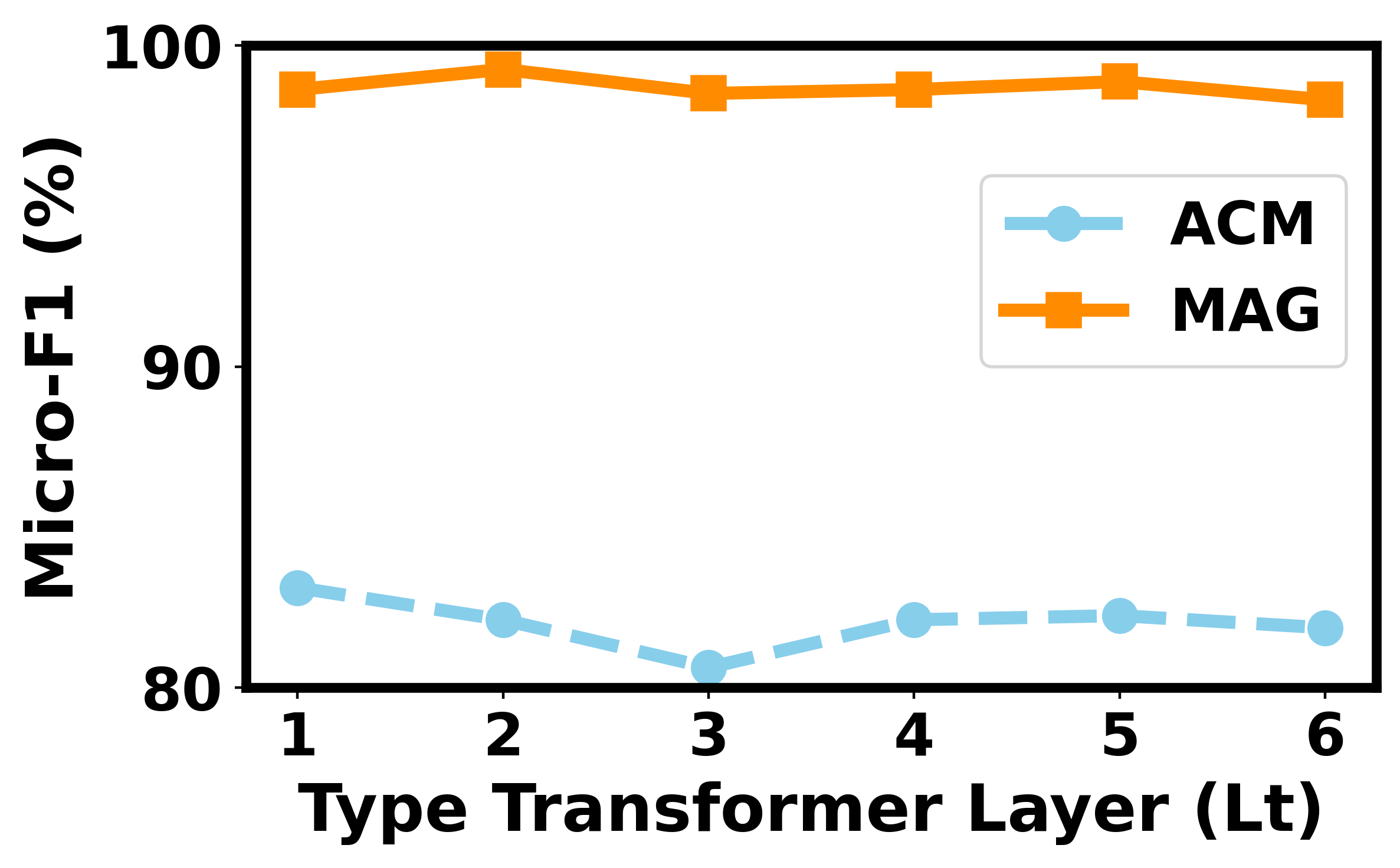}}
  \subfloat[Macro-F1]{\label{fig:nc_lt_ma}\includegraphics[width=0.2\textwidth]{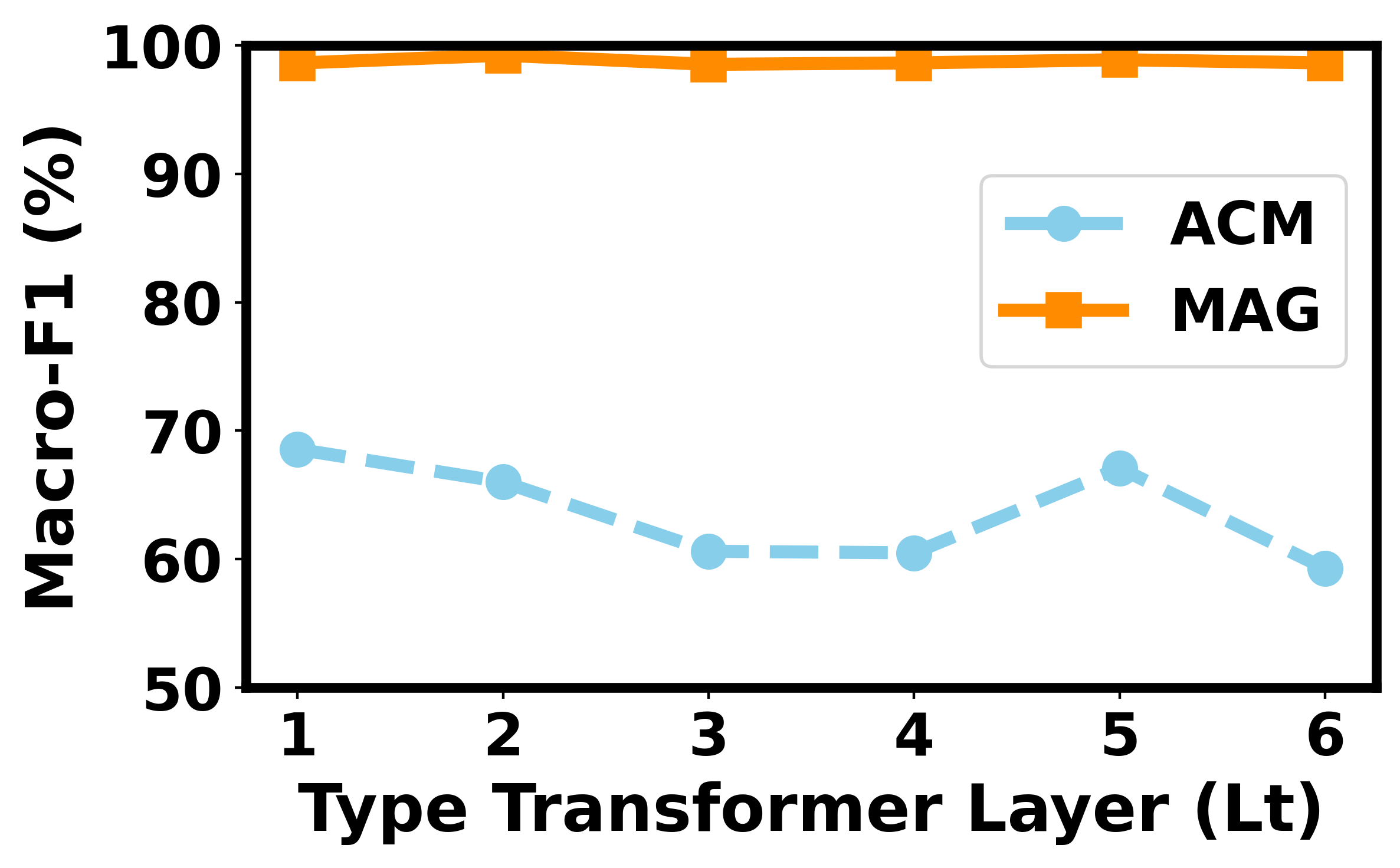}}
  \caption{{\zlcikm Type-level} transformer layer {\zqycikm number} study for node classification.}   
  \label{fig:cla_lt}  
\end{figure}

\noindent \textbf{Effect of Transformer Layer $Lr$ and $Lt$.} To estimate the sensitivity of Transformer layer, we vary the Ring-level Transformer layer $Lr$ and Type-level Transformer layer $Lt$ in $\{1,2,3,4,5,6\}$ while fixing other parameters fixed. For the Ring-level Transformer layer, as shown in Figure~\ref{fig:cla_lh}, the best results on both datasets are achieved with $Lr=4$. Additionally, we observe different trends in performance with increasing $Lr$ on both MAG and ACM datasets, which can be attributed to their unique characteristics and structural differences. For the MAG dataset, increasing $Lr$ initially improves performance by capturing more complex patterns. However, further increases in $Lr$ lead to a slight decline in performance, indicating that the model's capacity becomes too large, resulting in overfitting and over-smoothing. In contrast, on the ACM dataset, performance decreases initially with increasing $Lr$, but then improves as $Lr$ continues to increase, suggesting that the model begins to capture more relevant patterns as its capacity grows.

For the Type-level Transformer layer as shown in Figure~\ref{fig:cla_lt}, we observe similar results due {\zqycikm to} the same reasons. Besides, the best results on both datasets are achieved with different $Lt$ values, since different datasets exhibit distinct characteristics. }

\noindent \textbf{Effect of Embedding Size $d$.} We vary $d$ in $\{128, 256,$  $512, 1024, 2048\}$ to validate the impact of embedding size. Figure~\ref{fig:cla_d} reports the node classification results over {\zqycikm both} datasets. 
As observed, in most cases, model performance improves with increasing hidden dimension size, as a larger embedding size generally provides stronger representational power. However, it is interesting to discover that employing high-dimensional representations does not consistently yield optimal results.  For instance, the model achieves the optimal Micro-F1 and Macro-F1 when $d=128$ on ACM dataset and $d=512$ on MAG dataset, respectively. This indicates adopting a higher-dimensional representation does not guarantee the best performance across all scenarios.

\noindent \textbf{Effect of Ring Number $K$.} We range $K$ from $1$ to $10$ to analyze the effect of the number of rings, and the node classification results are illustrated in Figure~\ref{fig:cla_K}. As observed, the model achieves the best with different $K$ on different datasets, since various HINs display distinct neighborhood configurations. Besides, as $K$ increases, performance gradually improves across all datasets, followed by a slight decline observed with further increments. Though greater $K$ implies nodes consider a broader neighborhood, too large $K$ may cover the entire network, causing the node's neighborhood to include a significant amount of irrelevant information and even over-fitting.
  
\section{Conclusion}
\label{sec:conclusion}
In this paper, we study the HIN representation {\zqy learning} problem. 
{\qy To deal with it, we introduce an innovative {\zlcikm $(k,t)$-ring neighborhood structure} to extract neighbors for each node, {\zlcikm aiming to capture the differences between neighbors at distinct distances and with different types.}}
Based on this novel structure, we propose an effective HHGT model, seamlessly integrating a Type-level Transformer for aggregating nodes {\qy of} different types within each $k$-ring neighborhood, and a Ring-level Transformer for hierarchical aggregation across multiple $k$-ring neighborhoods. Experimental results on two real-world datasets demonstrate the advantages of our HHGT model across various downstream tasks.
\newpage  


\bibliographystyle{ACM-Reference-Format}
\bibliography{hhgt}

\appendix

\end{document}